\documentclass[review]{elsarticle}

\usepackage{lineno,hyperref}
\usepackage{graphicx}
\usepackage{enumerate}
\usepackage[small,bf,hang]{caption} 
\usepackage{subcaption}
\usepackage{amsmath}
\usepackage{amssymb}
\usepackage{rotating}
\usepackage{xcolor}		 	
\usepackage{verbatim}		
\usepackage{colortbl}
\usepackage{booktabs}
\usepackage{dcolumn}
\usepackage{expdlist}  
\usepackage{float}
\usepackage{url}
\usepackage{fullpage}
\usepackage{mathtools}
\usepackage{amsthm}
\usepackage{textcomp}
\usepackage{makecell}

\usepackage{longtable}
\usepackage{threeparttable}
\usepackage{threeparttablex}
\usepackage[ruled,vlined]{algorithm2e}
\usepackage{optidef}
\usepackage{IEEEtrantools}

\usepackage{mathptmx}   
\usepackage{pdfpages}
\usepackage{adjustbox}
\usepackage{array}
\usepackage{pifont}
\usepackage[onehalfspacing]{setspace}
\usepackage{cclicenses}  
\newcommand\blfootnote[1]{%
  \begingroup
  \renewcommand\thefootnote{}\footnote{#1}%
  \addtocounter{footnote}{-1}%
  \endgroup
}

\newcolumntype{R}[2]{%
    >{\adjustbox{angle=#1,lap=\width-(#2)}\bgroup}%
    l%
    <{\egroup}%
}
\newcommand*\rot{\multicolumn{1}{R{45}{1em}}}

\DeclarePairedDelimiter\ceil{\lceil}{\rceil}
\DeclareMathOperator*{\argmaxB}{argmax}

\newtheorem{definition}{Definition}
\newtheorem{theorem}{Theorem}

\SetKwFunction{Range}{range}

\SetKwProg{FnA}{Function}{}{end}\SetKwFunction{FRecursA}{makeTree}%
\SetKwProg{FnB}{Function}{}{end}\SetKwFunction{FRecursB}{obtainInitialSpanningTree}%
\SetKwProg{FnC}{Function}{}{end}\SetKwFunction{FRecursC}{computeEarliestTime}%
\SetAlgoLongEnd

\newcommand{\cmark}{\ding{51}}%
\newcommand{\xmark}{\ding{55}}%

\SetKwFunction{Range}{range}
\SetKwFor{While}{while}{:}{fintq}%
\SetKw{KwTo}{in}\SetKwFor{For}{for}{\string:}{}%

\newcommand{\forcondC}{$b$  \KwTo{$\{0,1,\ldots,n-1\}$}}
\newcommand{\forcondD}{$k$  \KwTo{$\{m-1,m-2,\ldots,0\}$}}
\modulolinenumbers[5]
\newcommand{\blue}[1]{\textcolor{black}{#1}}

\journal{Computers \& Operations Research}

\begin{document}

\begin{frontmatter}

\title{Exploring search space trees using an adapted version of Monte Carlo tree search for combinatorial optimization problems}


\author[KULAK]{Jorik Jooken\corref{mycorrespondingauthor}}
\ead{jorik.jooken@kuleuven.be}
\cortext[mycorrespondingauthor]{Corresponding author}

\author[KULAK,UGENT,FLANDERSMAKE]{Pieter Leyman}
\ead{pieter.leyman@ugent.be;pieter.leyman@kuleuven.be}

\author[CODESGENT]{Tony Wauters}
\ead{tony.wauters@kuleuven.be}

\author[KULAK]{Patrick De Causmaecker}
\ead{patrick.decausmaecker@kuleuven.be}

\address[KULAK]{Department of Computer Science, KU Leuven Kulak, Etienne Sabbelaan 53, 8500 Kortrijk, Belgium}
\address[UGENT]{Department of Industrial Systems Engineering and Production Design, Ghent University, Technologiepark Zwijnaarde 46, 9052 Zwijnaarde, Belgium}
\address[FLANDERSMAKE]{Industrial Systems Engineering, Flanders Make@UGent, Sint-Martens-Latemlaan 2B, 8500 Kortrijk, Belgium}
\address[CODESGENT]{Department of Computer Science, CODeS, KU Leuven, Gebroeders De Smetstraat 1, 9000 Ghent, Belgium}




\begin{abstract}
In this article we propose a heuristic algorithm to explore search space trees associated with instances of combinatorial optimization problems. The algorithm is based on Monte Carlo tree search, a popular algorithm in game playing that is used to explore game trees and represents the state-of-the-art algorithm for a number of games. Several enhancements to Monte Carlo tree search are proposed that make the algorithm more suitable in a combinatorial optimization context. These enhancements exploit the combinatorial structure of the problem and aim to efficiently explore the search space tree by pruning subtrees, using a heuristic simulation policy, reducing the domains of variables by eliminating dominated value assignments and using a beam width. The algorithm was implemented with its components specifically tailored to two combinatorial optimization problems: the quay crane scheduling problem with non-crossing constraints and the 0-1 knapsack problem. For the first problem our algorithm surpasses the state-of-the-art results and several new best solutions are found for a benchmark set of instances. For the second problem our algorithm typically produces near-optimal solutions that are slightly worse than the state-of-the-art results, but it needs only a small fraction of the time to do so. These results indicate that the algorithm is competitive with the state-of-the-art for two entirely different combinatorial optimization problems.
\end{abstract}


\end{frontmatter}

\section{Introduction}
	\label{intro}
	\blfootnote{This work is licensed under a Creative Commons Attribution-NonCommercial-NoDerivatives 4.0 International License \cc \byncnd}\blfootnote{DOI: \url{https://doi.org/10.1016/j.cor.2022.106070}} The operations research community has shown an increasing interest in approaches that use machine learning to solve combinatorial optimization problems. One of the main advantages of these approaches is that they can help to improve certain human-made algorithmic decisions. Some of these approaches rely almost exclusively on machine learning (e.g. \cite{Vinyals:2015,Bello:2017} and \cite{Khalil:2017}), whereas other approaches combine machine learning with ideas found in more traditional approaches to solve combinatorial optimization problems (e.g. \cite{Hottung:2016} and \cite{Karapetyan:2017}). In this article, we follow the latter setting and we propose several adaptations for the Monte Carlo tree search algorithm \cite{Coulom:2006}, originally used in game playing, in order to efficiently explore the search space associated with a problem instance of a combinatorial optimization problem.

Solving such a problem instance boils down to finding an assignment of values to the decision variables such that all constraints are met and the objective function is minimized (without loss of generality). In case the domains of all decision variables are finite sets, the set of all feasible solutions can be represented by means of a rooted tree that represents the complete search space. The search space tree, and thus the set of all solutions, is often too big to exhaustively enumerate in a reasonable time. Nevertheless, there are many well-known techniques that still make use of this tree to solve the optimization problem. Each technique copes with this issue of a large search space in its own way. An example of a technique that can be used to explore a search space tree is branch-and-bound, which can skip exploring entire subtrees by proving that the set of solutions in a subtree are all worse than another solution. Other techniques cope with this issue by sacrificing the optimality guarantee of the algorithm in \blue{return} for a time gain. These heuristic techniques only explore a subset of all feasible solutions and choose the best one in this subset (e.g. beam search). Finally, there are techniques that also only explore a subset of all solutions, but they do not explicitly use the search space tree to do so (e.g. several metaheuristics and algorithms based on local search).

	The main contributions of this article are twofold. First, the article presents a heuristic algorithm to explore search space trees that is based on Monte Carlo tree search, a popular reinforcement learning algorithm for game playing \cite{Sutton:2018,Coulom:2006}. We show that this algorithm can be modified in several ways to combine machine learning with ideas found in more traditional approaches for solving combinatorial optimization problems. These modifications all exploit the combinatorial structure of the problem to efficiently explore the search space. The proposed algorithm is able to prune large parts of the search space tree by using bounds on the objective function value. Furthermore, it is possible to integrate heuristic, problem specific information into the algorithm by means of a heuristic simulation policy and the domains of the decision variables can be reduced by eliminating dominated value assignments (more details will follow later in Section \ref{MCTSCombinatorialOptimization}). The algorithm has the ability to automatically learn how to navigate through the search space tree, just like regular Monte Carlo tree search for game playing can learn this for game trees. To speed up this learning process, and hence to keep learning in a very big search space manageable, the algorithm employs the idea of using a beam width (see for example \cite{Baier:2012}). By doing this, the search space is explicitly shrunk by removing possible solutions that do not look promising. The current article is the first study that uses all of these modifications at the same time. We will show that these modifications greatly improve the results in comparison with the non-modified version and that omitting any of the modifications yields (sometimes drastically) worse results. Second, the algorithm was empirically validated on two case studies using a set of extensive experiments requiring around 1,050 CPU-hours. For each case study, we instantiated the proposed algorithm by tailoring the problem specific components of the algorithm (i.e. we used a custom heuristic simulation policy, custom bounds on the objective function values and custom dominance rules for each case study). The first case study concentrates on the quay crane scheduling problem with non-crossing constraints \cite{Lee:2010,Santini:2014}, which is classified in literature as $[1D||C_{Max}]$ according to the classification scheme proposed in \cite{Boysen:2017}. For this case study, our algorithm surpasses the state-of-the-art results and several new best solutions are found. The second case study concentrates on one of the most studied combinatorial optimization problems: the 0-1 knapsack problem \cite{Kellerer:2004}. For this case study our algorithm typically produces near-optimal solutions that are slightly worse than the state-of-the-art results, but it needs only a small fraction of the time to do so.

	The rest of this article is organized as follows: in Section \ref{MCTSGamePlaying} the Monte Carlo tree search algorithm for game playing is explained. An overview of the relevant literature for this article is given in Section \ref{literatureOverview}, followed by the adaptations that we propose for Monte Carlo tree search in the context of combinatorial optimization (Section \ref{MCTSCombinatorialOptimization}). Next, in Section \ref{caseStudy} the quay crane scheduling problem with non-crossing constraints is introduced and the different problem specific components of the proposed algorithm are concretely demonstrated for this problem. Similarly, the same structure is followed in Section \ref{caseStudyB}, but this time for the 0-1 knapsack problem. The computational results for both case studies are discussed in Section \ref{computationalResults}. Finally, a conclusion and possible ideas for future work are given in Section \ref{conclusions}.
	\section{Monte Carlo tree search for game playing}
	\label{MCTSGamePlaying}
	
	Monte Carlo tree search is a heuristic search algorithm that is popular in game playing. Variants of this algorithm have been successfully applied to a variety of games (e.g. Havannah \cite{Dugueperoux:2016}, Amazons \cite{Lorentz:2008}, Lines of Action \cite{Winands:2010}, Hex \cite{Arneson:2010}, Go \cite{Silver:2016}, chess and Shogi \cite{Silver:2017}) and represent the state-of-the-art approach for many of them. The algorithm is capable of learning to play promising moves in a turn-based game. It does this by iteratively improving an estimate of how good a move is by using Monte Carlo sampling. Initially, these estimates are highly uncertain, but as more samples are collected, the estimates become more accurate. \blue{Thus}, the algorithm's performance improves over time and it effectively learns how to play the game. The algorithm performs Monte Carlo sampling on a game tree to estimate the quality of the moves. A game tree is a tree in which each node represents a possible state of the game and the edges between the nodes represent the possibility to go from one state to another state by playing a move. The root node of the game tree represents the initial state of the game and leaf nodes correspond to final states (i.e. for two-player games a leaf node represents a state in which the game was won by one of the players). A detailed explanation of how the algorithm operates on the game tree follows next.
	
	The algorithm keeps track of a part of the game tree, which we will denote as the \textit{partial game tree} further on. The partial game tree is a subtree of the whole game tree with the same root node. Initially, the partial game tree only consists of a single node: the root node of the game tree. The partial game tree is iteratively grown until a certain computational budget is exhausted (e.g. a time limit or a number of iterations). In every iteration, one game is played until the end. Playing one game until the end corresponds to walking along a path from the root of the game tree to a leaf node. Every iteration consists of four phases:

\begin{enumerate}[\bfseries Ph{a}se 1:]
\item \textbf{Selection}. The algorithm starts in the root of the game tree and will walk further down until it reaches a node of the game tree that does not yet belong to the partial game tree. When the algorithm walks down the game tree, it has to consecutively choose which child it will descend to. To make this decision, the algorithm tries to find a good balance between exploitation (playing moves that seem very good) and exploration (playing moves that have not been played very often in previous iterations in order to get a better estimate of how good the move is). To do this, it treats the problem of selecting which child to descend to for every node as a multi-armed bandit problem \cite{Lai:1985}. In a multi-armed bandit problem there is a player and there are $k$ levers $l_1, l_2, ..., l_k$. With every lever $l_i$ we associate a probability distribution $P_i$ such that if lever $l_i$ is pulled, the player receives a random reward sampled from $P_i$. These probability distributions are unknown to the player. The player will play multiple rounds and in each round the player will pull exactly one lever. The objective of the multi-armed bandit problem is to find a policy that determines which lever the player should pull in every round in order to maximize the expected sum of the received rewards. In the context of Monte Carlo tree search, this problem has to be solved every time a child node has to be selected from a given parent node. Hence, the levers in the multi-armed bandit problem correspond to the children of a node of the game tree in Monte Carlo tree search. The rewards correspond to 0 or 1 for a loss and win, respectively. Kocsis and Szepesv\'{a}ri solve this problem in their UCT algorithm \cite{Kocsis:2006} for Monte Carlo tree search by selecting the child for which the expression:
\begin{equation}
\label{argmaxExpression}
\frac{numberWins(parent,child)}{numberVisits(child)}+\sqrt{\frac{2 \cdot \ln{(numberVisits(parent))}}{numberVisits(child)}}
\end{equation}
is maximized\footnote{In \cite{Kocsis:2006}, it was mentioned that the algorithm has to be implemented such that division by zero is avoided. Our approach to this corner case will be explained later in Section \ref{MCTSCombinatorialOptimization}.}. Here, $numberVisits(v)$ denotes the number of iterations in which node $v$ was visited (belonged to a generated path) and $numberWins(v,w)$ denotes the number of iterations in which the edge from parent node $v$ to child node $w$ was followed and the game was won (by the player whose turn it is in the game state associated with node $v$). By maximizing this expression, the algorithm attempts to find a good balance between exploitation and exploration. The first term of this expression is the average win ratio and will be high for good moves, whereas the second term of this expression will be high for moves that have not been explored very often. It has been shown in \cite{Auer:2002} that this policy is optimal in the sense that the expected sum of rewards is asymptotically (when the number of rounds goes to infinity) as high as possible over all possible policies.
\item \textbf{Expansion}. The algorithm has arrived in a node which it has never visited before. For this node there is no information available from previous iterations to assess how good the possible moves are. The partial game tree is extended by adding this node to the tree.
\item \textbf{Simulation}. The algorithm will further walk down the game tree until a leaf node (a terminal game state) is reached. There are many possible policies to walk down the game tree. However, in the most basic version of Monte Carlo tree search, the policy recursively selects uniformly at random one of the available children of the current node until a leaf node is reached.
\item \textbf{Backpropagation}. The outcome of the game is evaluated by inspecting the game state associated with the leaf node. This information is backpropagated to all nodes $v$ of the constructed path that belong to the partial game tree by updating $numberVisits(v)$ and $numberWins(v,w)$ (i.e. $numberVisits(v)$ is incremented and $numberWins(v,w)$ is incremented if and only if the player whose turn it is in the game state associated with $v$ has won the game). The effect of this is that moves that were played during this iteration by the winner of the game will be more likely to be played again in future iterations.
\end{enumerate}

\begin{figure*}[h!]
	\centering
  \includegraphics[width=0.6\textwidth]{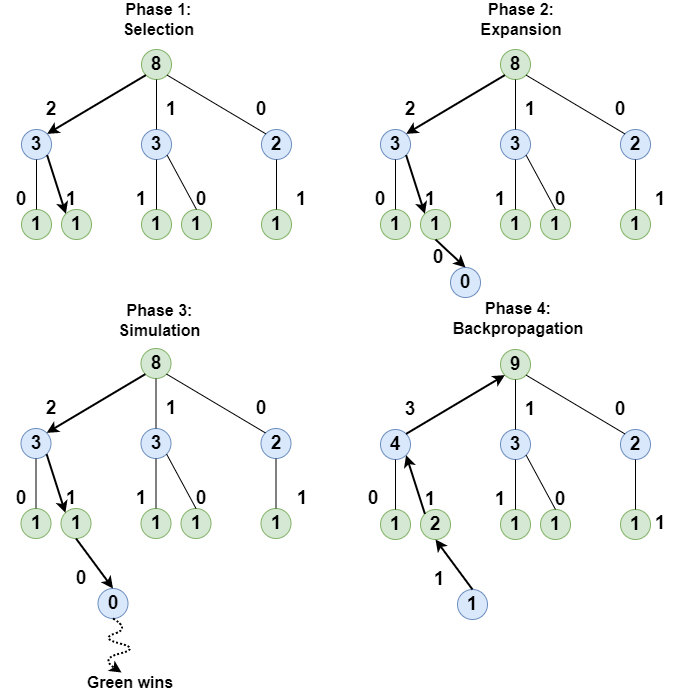}
\caption{The four phases of an iteration of Monte Carlo tree search. Here, $numberVisits(v)$  and $numberWins(v,w)$ are indicated for every node and edge, respectively.}
\label{fourPhases}       
\end{figure*}

These four phases are depicted in Figure \ref{fourPhases}. A partial game tree is shown for a turn based two-player game with a green player and a blue player \blue{(we refer readers of the physical journal to the online version of the paper, where colors are available)}. The colours of the nodes represent whose turn it is. For every node and edge, respectively $numberVisits(v)$ and $numberWins(v,w)$ are indicated. The moves that are played in Phase 1 are indicated by the arrows. In Phase 2, the partial game tree is extended and consists of ten nodes after the extension. Next, the algorithm walks further down the game tree in Phase 3 and ends in a final state in which the green player has won the game. Finally, the outcome of this game is backpropagated in Phase 4 by updating $numberVisits(v)$ and $numberWins(v,w)$ for all nodes and edges of the generated path in the partial game tree.

	\section{Literature overview}
	\label{literatureOverview}

	The algorithm that we propose in this article is a heuristic search algorithm based on Monte Carlo tree search that is enhanced in several ways in the context of solving combinatorial optimization problems. This algorithm was validated on two different case studies: the quay crane scheduling problem with non-crossing constraints (more specifically, the version that is known in literature as $[1D||C_{Max}]$) and the 0-1 knapsack problem. Hence, the relevant literature for this article comes from several different contexts. In this section, we will give an overview of related work concerning Monte Carlo tree search and we will also give an overview of the literature concerning the two problems from the case studies. The overviews are not meant to be exhaustive, but rather discuss the most important results relevant for this article. We refer the interested reader to \cite{Browne:2012}, \cite{Bierwirth:2015} and \cite{Kellerer:2004} for a more complete overview of respectively Monte Carlo tree search, the quay crane scheduling problem and the 0-1 knapsack problem.

	\subsection{Monte Carlo tree search}
	In this subsection we will discuss Monte Carlo tree search from the perspective of adaptations that have been proposed for the different phases of the algorithm as well as other contexts than game playing in which adaptations of Monte Carlo tree search have been used. Chaslot et al. propose \textit{progressive bias} \cite{Chaslot:2008} to improve Monte Carlo tree search by changing expression (\ref{argmaxExpression}) that is used to select children in Phase 2 of the algorithm. They add a weighted term to the expression that enables them to embed domain specific knowledge into the expression. In this term, a heuristic value is calculated for every possible child (moves that seem to be better according to the heuristic, get a higher value). The weight of this term (and hence the bias) becomes smaller after every iteration and goes to zero as the number of iterations goes to infinity. Winands et al. \cite{Winands:2009} further build upon this idea. They use this adapted expression only after the nodes have been visited a fixed number of times (and use the same policy as in the simulation phase otherwise) in order to avoid spending too much time on computing the heuristic values. Gelly et al. \cite{Gelly:2007} embed domain specific knowledge into the selection phase of Monte Carlo tree search by integrating heuristic values for the moves that were calculated offline. Instead of initialising the values $numberVisits(v)$ and $numberWins(v,w)$ by 0, the heuristic values are used to appropriately initialise them, leading to a hot-start of the algorithm. Li et al. \cite{Li:2021} propose a different \blue{static} selection policy that optimally allocates a limited computing budget to maximize a lower bound on the probability of correctly selecting the best action at each node (whereas many papers focus on maximizing the cumulative rewards). \blue{Zhang et al. \cite{Zhang:2022} recently proposed a closely related selection policy, which is dynamic instead of static.} Baier and Winands \cite{Baier:2012} proposed Beam Monte Carlo tree search in which the number of traversed nodes at a given depth is restricted as soon as a certain number of iterations has been performed at that depth by only keeping the $w$ most visited nodes. This idea is very similar to one of the modifications that we proposed in the context of Monte Carlo tree search for combinatorial optimization in this article (subsection \ref{beamWidthSubsection}), except for two differences. We keep the $w$ most promising nodes according to the average objective function value instead of number of visits and the beam width restriction is applied for every level after a certain time has passed depending on the depth of the tree instead of a fixed number of iterations.

	Winands et al. \cite{Winands:2008} focus on improving the simulation in Phase 3 of the algorithm. Instead of only evaluating the final game state as a winning state or losing state, they also attempt to prove for intermediate states the game-theoretic value (i.e. deciding whether a state is a proven loss or win). In case the game-theoretic value of a node can indeed be proven, the iteration is stopped and this value is backpropagated such that its outcome can be used again for future iterations. Drake et al. \cite{Drake:2007} attempt to improve the simulation by using domain specific knowledge. They use \textit{heavy playouts}, in which a heuristic is used that is specifically tailored to the game under consideration as opposed to the default random move policy. Another approach is followed by AlphaGo \cite{Silver:2016}, in which an artificial neural network is trained to predict the outcome of a game. Here, the game is not simulated in the simulation phase, but instead the outcome of the game is predicted and this value is used for backpropagation.

	Gelly et al. \cite{Gelly:2011} focus on improving Phase 4 of Monte Carlo tree search. In some games, all permutations of a set of moves lead to the same game state. To be able to share information amongst different states, they use the outcome of games in which an action $a$ was played not only to update the node in which action $a$ was played, but also all nodes in the subtree that has this node as root node. This provides a biased, but rapid value estimate. In their MC-RAVE algorithm \cite{Gelly:2011}, this bias decreases after many iterations by introducing a weight for the biased term that goes to zero as the number of iterations increases. Another attempt to improve the backpropagation phase can be found in \cite{Xie:2009}. Because the estimates of quality of the moves get better after every iteration, the estimates get more and more reliable. To account for this, a weight factor is introduced that gives a higher weight to iterations that are deemed more reliable.

	Adaptations of Monte Carlo tree search have also been used in other contexts than game playing. It has been used in the context of constraint satisfaction problems by Previti et al. to obtain good results for structured instances of the SAT problem \cite{Previti:2011}. Satomi et al. \cite{Satomi:2011} adapt Monte Carlo tree search to solve large-scale quantified constraint satisfaction problems in real-time. Perez et al. \cite{Perez:2013} use Monte Carlo tree search to build a controller for the physical travelling salesman problem (a real-time game). Two other surprising application domains for Monte Carlo tree search can be found in the work of Tanabe et al. where Monte Carlo tree search was used in a security context to detect \textit{wolf attacks} \cite{Tanabe:2009} and the work of Dieb et al. \cite{Dieb:2017} in which the task of automatic complex material design is studied. 

\blue{It is also worth mentioning} other uses of Monte Carlo tree search in a combinatorial optimization context and highlighting the most important differences with respect to the present article. In the work of Sabar et al. \cite{Sabar:2015} Monte Carlo tree search is used in a hyper-heuristic \cite{Ross:2005} as a high level selection strategy to select a low-level heuristic. The most important difference with the present article is that the tree on which the algorithm operates is completely different. In the tree on which the algorithm operates in the work of Sabar et al., every node represents a low-level heuristic (e.g. a specific perturbation operator) and every path in the tree corresponds to a sequence of low-level heuristics. The goal is to find a sequence of low-level heuristics that can be applied to an initial starting solution in order to obtain a good solution. In the present article on the contrary, the tree on which the algorithm operates represents the search space of all possible assignments of values to decision variables and the goal is to find a leaf node (representing a solution in which all variables have been instantiated) for which the objective function value is as small as possible (without loss of generality). Another use of Monte Carlo tree search was reported in the work of Sabharwal et al. \cite{Sabharwal:2012}, in which Monte Carlo tree search was used to guide the node selection heuristic for the Mixed Integer Programming (MIP) solver CPLEX to select which node to expand. Again, the tree on which the algorithm operates is very different from the search space tree in the present article. Every node of the tree represents the solution of a linear programming relaxation of a mixed integer program and every edge corresponds to a branching decision for a non-integer variable. Furthermore, in the work of Sabharwal et al., no solution is constructed in the simulation phase of the algorithm, but instead a bound on the objective function value is used for the backpropagation phase. 

\blue{We now turn our attention to} literature in which the tree on which Monte Carlo tree search operates is the search space tree associated with a problem instance (as is also the case in the current article). Chaslot et al. \cite{Chaslot:2006} use Monte Carlo tree search to solve production management problems and obtain better solutions than an evolutionary planning heuristic. Liu et al. \cite{Liu:2019} propose a smart directed acyclic graph scheduling algorithm based on Monte Carlo tree search, in which suboptimal parts of the search space are pruned using a custom dual bound. Rosin \cite{Rosin:2011} proposes the Nested Rollout Policy Adaptation algorithm (a variant of Monte Carlo tree search in which the rollout policy is gradually adapted). They use this algorithm to produce good solutions for instances of Crossword Puzzle Construction and Morpion Solitaire. A more classical problem is tackled by Grelier et al. \cite{Grelier:2022}. They combine several dedicated heuristics for the weighted vertex covering problem with Monte Carlo tree search and provide empirical evidence for the advantages and disadvantages of the different algorithmic variants. Similarly, Cazenave et al. \cite{Cazenave:2020} use the Nested Rollout Policy Adaptation algorithm and Nested Monte Carlo Search \cite{Cazenave:2009} (another Monte Carlo tree search variant) for solving the graph colouring problem. Although the results obtained by \cite{Cazenave:2020} do not surpass the state-of-the-art, they are almost as good (or equally good) for several problem instances. Runarsson et al. \cite{Runarsson:2012} compare the pilot method with Monte Carlo tree search to solve the Job Shop Scheduling problem \cite{VanLaarhoven:1992} and conclude that the latter algorithm often outperforms the former one. Edelkamp et al. \cite{Edelkamp:2016} use the Nested Rollout Policy Adaptation algorithm to solve several variations of classical problems in logistics: a vehicle routing problem, a motion planning problem and a container packing problem. For each problem, they report a promising solution for one problem instance.

As we can see from this subsection, several variants of Monte Carlo tree search have been proposed throughout the years. Given that the literature on Monte Carlo tree search is very large, it is not surprising to see that some of the enhancements from the current article (see Section \ref{MCTSCombinatorialOptimization}) also appear in a limited number of other articles. However, they typically appear in isolation, whereas the current article is the first study that uses all enhancements at the same time. As we will later show in Section \ref{computationalResults}, the effect of omitting one or more of these enhancements can be quite big and the best results are obtained when all the enhancements are used. Using all enhancements at the same time is especially important for the most challenging problem instances.

	\subsection{The quay crane scheduling problem with non-crossing constraints}
	Quay cranes are used in ports for the loading and unloading of the bays of a container vessel. In order to guarantee an efficient execution, an important problem that arises in this context consists of finding a schedule for the quay cranes such that some quality criteria (e.g. the makespan or crane utilization rate) are optimized. The quay crane scheduling problem was first formulated by Daganzo in 1,989 \cite{Daganzo:1989}. They proposed an MIP to solve the quay crane scheduling problem, which was later improved in 1,990 by Peterkofsky et al. \cite{Peterkofsky:1990}. These two seminal papers attracted a lot of attention to this problem and led to a vast amount of literature that is concerned with modeling and solving it. The models in literature all describe the same concept of quay crane scheduling, but differ significantly in several aspects (e.g. the level of detail of the tasks, the objective function to be optimized, the given problem parameters and constraints). From the point of view of combinatorial optimization, this makes it difficult to compare the different models, because a problem instance for one formulation of the problem is not necessarily a valid problem instance for another formulation. Recently, a classification scheme was proposed by Boysen et al. \cite{Boysen:2017} to group the different formulations in different classes according to three criteria: the terminal layout, the characteristics of the container moves and the objective function. The problem that we study in this article  is classified with the name $[1D||C_{Max}]$ according to the classification scheme of Boysen et al. \cite{Boysen:2017}. In this problem the terminal layout can be considered to be a one-dimensional line along which both the quay cranes and bays are located. There are no special characteristics of the container moves and the objective function that has to be minimized is the makespan (the last completion time of a bay). For this reason, we will restrict the rest of this subsection to an overview of related literature for the problem $[1D||C_{Max}]$ instead of general \blue{quay crane scheduling} literature.

	Zhu and Lim \cite{Zhu:2006} model the quay crane scheduling problem as an integer program, which they solve by a branch-and-bound algorithm and also propose a simulated annealing algorithm for larger instances. This approach is significantly improved by Lim et al. \cite{Lim:2007} by making the crucial observation that instead of directly focusing on the schedule itself, it suffices to focus on the allocation of quay cranes to bays. More specifically, they prove that there exists an optimal schedule for their problem such that the cranes move in only one direction and this schedule can be derived from the allocation of quay cranes to bays. They also prove that the quay crane scheduling problem with non-crossing constraints is NP-hard and propose an approximation algorithm which guarantees an approximation factor of 2. Lee et al. \cite{Lee:2008} also propose an MIP for small problem instances and a genetic algorithm for larger problem instances. Lee and Chen \cite{Lee:2010} later realised that the models from \cite{Zhu:2006}, \cite{Lim:2007} and \cite{Lee:2008} can lead to solutions in which two quay cranes would have to occupy the same position at the same time and they propose an MIP that mitigates this deficiency (the specific quay crane scheduling problem that is studied in the present article follows the model proposed in \cite{Lee:2010}). They also propose a heuristic two-stage algorithm, which they call enhanced best partition (EBP). In the first stage of the algorithm, the bays are partitioned into consecutive areas such that the resulting solution guarantees an approximation factor of 2. In the second stage, this solution is improved by merging adjacent areas into one area until no further improvements can be found. Lee and Wang \cite{Lee:2010b} study an extension of $[1D||C_{Max}]$: the integrated discrete berth allocation and quay crane scheduling problem and propose an MIP for small problem instances and a genetic algorithm for larger instances. Another extension is studied in \cite{Tang:2014}, where the quay crane scheduling problem is integrated into the truck scheduling problem. They propose both an MIP and an algorithm based on Particle Swarm Optimization to solve the problem. Santini et al. \cite{Santini:2014} improve the MIP model of Lee and Chen \cite{Lee:2010} by introducing a family of valid inequalities such that the set of feasible solutions does not change, but the efficiency of the MIP solver CPLEX is greatly affected. Finally, Zhang et al. \cite{Zhang:2017} propose an approximation algorithm for $[1D||C_{Max}]$ and prove that their approximation factor is smaller than 2 (the best known approximation ratio until then). The achieved approximation ratio is $2-\frac{2}{m+1}$, where $m$ denotes the number of quay cranes.

	\subsection{The 0-1 knapsack problem}
	The 0-1 knapsack problem is a classical problem which has already received attention for several decades. In this problem, the goal is to select several items from a given set of $n$ items to include in a knapsack with limited capacity $c$ such that the sum of the weights of the selected items does not exceed $c$ and the sum of their profits is maximized. The problem is NP-hard \cite{Garey:1979}, but it can be solved in pseudo-polynomial time (see e.g. \cite{Dantzig:1957} for a solution with a worst-case time complexity of $O(n \cdot c)$) and several fully polynomial time approximation schemes exist (see e.g. \cite{Lawler:1979}, \cite{Magazine:1981}, \cite{Kellerer:1999} and \cite{Kellerer:2004b}). This is mainly interesting from a theoretical point of view, but it does not immediately give rise to algorithms with a good practical performance \cite{Kellerer:2004}. However, such practical algorithms do exist. Many of these algorithms are based on ordering the items according to the ratio between the profit and the weight of an item. This ordering can be used to efficiently solve the linear programming relaxation of the 0-1 knapsack problem and this gives rise to a well known upper bound by Dantzig \cite{Dantzig:1957}. This ordering is also used by a simple greedy heuristic that selects items in this order as long as the knapsack capacity is not exceeded. The resulting solution often differs from the optimal solution by only a few items whose profit-weight ratio is close to a certain distinguished item called the split item \cite{Kellerer:2004}. Martello et al. \cite{Martello:1988} proposed an algorithm that first determines a small subset of items, which is called the core, such that for items in the core it is usually difficult to decide whether they will belong to an optimal solution, whereas for items outside of the core this decision is usually easy. Next, the algorithm optimally selects items from the core and combines these with items outside of the core. A drawback of this algorithm is that the core size is fixed, but it is very difficult to know beforehand which core size will be appropriate. This drawback was tackled by Pisinger \cite{Pisinger:1995}, who proposed a branch-and-bound algorithm where the core is adaptively extended each time the algorithm reaches the border of the core. This was later improved even further by Pisinger \cite{Pisinger:1997} by handling the core enumeration process more efficiently. Finally, Martello et al. \cite{Martello:1999} proposed a combination of all the previous techniques, which resulted in their famous Combo algorithm. Although this algorithm was invented more than twenty years ago, it still represents the current state-of-the-art (see e.g. \cite{Buther:2012}, \cite{Monaci:2013}, \cite{Pisinger:2017} and \cite{Huerta:2020} for relatively recent articles that support this claim). It is able to exactly solve many problem instances containing several thousands of items in a matter of (milli)seconds. In an attempt to find the most challenging knapsack problem instances, Pisinger \cite{Pisinger:2005} has proposed 15 different classes of instances. Recently, another class of very hard problem instances was proposed in \cite{Jooken:2022}. Later in Section \ref{computationalResults}, we will focus in more detail on the most difficult class of problem instances proposed in \cite{Pisinger:2005} (strongly correlated spanner instances) and the class proposed in \cite{Jooken:2022} (noisy multi-group exponential problem instances).

	\section{Monte Carlo tree search for combinatorial optimization}
	\label{MCTSCombinatorialOptimization}	
	We propose an adapted version of Monte Carlo tree search for game playing (see Section \ref{MCTSGamePlaying}) that is more suitable for solving a combinatorial optimization problem. We assume that all decision variables are discrete, although in some cases the algorithm remains applicable even when the decision variables are continuous, as will be demonstrated in Section \ref{caseStudy}. By leveraging the combinatorial structure of a problem, the algorithm can be enhanced in several ways, which are discussed in this section. Later in Section \ref{caseStudy} and \ref{caseStudyB}, we will illustrate these enhancements by tailoring them to the specific combinatorial optimization problems of the two case studies. \blue{In this section we will employ the standard terminology used for combinatorial optimization problems and we refer readers who are unfamiliar with this terminology to \cite{Papadimitriou:1998}.}

	The algorithm will operate on the search space tree associated with a problem instance. Every node in this search space tree represents a partial solution (i.e. a solution for which at least one decision variable has not yet been assigned a value) except for the leaf nodes, which represent solutions in which all decision variables have been instantiated. The root of the tree corresponds to a partial solution in which no decision variable has been instantiated yet. Every edge represents the assignment of a value to a decision variable and all the edges on the same level of the tree correspond to value assignments to the same decision variable. Here, we assume that a natural ordering of the decision variables exists, which can often be obtained by modeling the optimization problem as making a natural sequence of decisions (note that the ordering of the decision variables often follows directly from the other components of the algorithm, as we will show later). We refer the interested reader to \cite{Morrison:2016} for an overview of heuristics that can be employed in case such a natural ordering does not exist. Every path from the root of the tree to a leaf node contains every decision variable exactly once. With this search space representation, solving the optimization problem now corresponds to finding a path from the root to a leaf node such that the associated solution is feasible and its objective function value is as small as possible. With this in mind, the algorithm will keep track of the best found feasible solution and the final result of the algorithm is the best solution found after all iterations have been executed. 

\begin{figure*}[h!]
	\centering
  \includegraphics[width=0.6\textwidth]{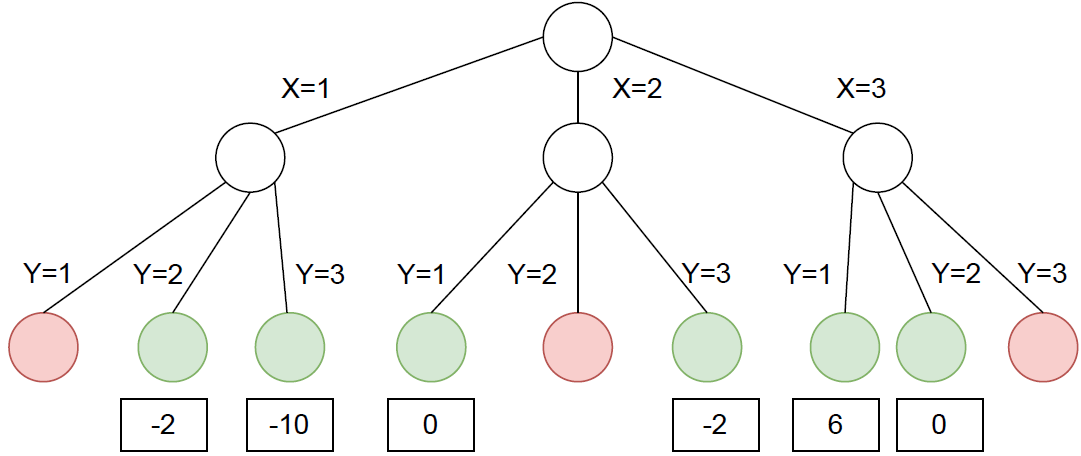}
\caption{An example of a search space tree for a toy problem instance}
\label{searchSpaceTreeFigure}       
\end{figure*}

An example of a search space tree is given in Figure \ref{searchSpaceTreeFigure}. For this problem instance, there are two decision variables $X$ and $Y$ that both have the same domain $\{1,2,3\}$. The search space tree is associated with a toy problem instance with objective function $(X-Y)^3-|X-Y|$ (to be minimized) and a single constraint $X \neq Y$. \blue{The search space tree begins by enumerating the domain of $X$ on the first level, followed by an enumeration of the domain of $Y$ on the second level. The root node corresponds to a partial solution in which no decision variable has been instantiated yet. This node has three children, each of which corresponds to a partial solution where the decision variable $X$ has been instantiated. The leaf nodes of the tree correspond to solutions (either feasible or infeasible) in which all decision variables have been instantiated. These} solutions are marked in green and red, respectively, with the optimal feasible solution having $X=1$, $Y=3$ and an objective function value of -10.

	The algorithm from the previous section was designed for game playing, but in this article the scope is combinatorial optimization. The similarity between these two is that there is the concept of a game tree in game playing and the concept of a search space tree in combinatorial optimization. Finding a solution for a problem instance of a combinatorial optimization problem can also be thought of as a kind of game in which the moves correspond to assigning a value to a decision variable. However, there are certain important differences between game playing and combinatorial optimization and these differences are the reason for the changes that we propose to the Monte Carlo tree search algorithm from the previous section. These differences are as follows:

\begin{description}
  \item[$\bullet$] First, there are typically only at most three possible outcomes for a game (win, loss and sometimes tie), whereas the objective function value of a problem instance of a combinatorial optimization problem can be any real number.
  \item[$\bullet$]  Second, the possible moves for a game are determined by the rules of the game and thus cannot be freely chosen, but the possible values that can be assigned to the decision variables for a problem instance of a combinatorial optimization problem depend on what these decision variables are (i.e. how the problem is modeled). Note that the same combinatorial optimization problem can often be modeled in several ways \blue{such that each one has} different decision variables, leading to very important consequences regarding for example the strength of bounds that can be derived from the model or the size of the search space.
  \item[$\bullet$] Third, there are several games that start from a fixed game state (e.g. chess and Go) and hence there is precisely one game tree associated with every game. For combinatorial optimization, however, there is one search space tree associated with every problem instance (as opposed to one search space tree for every problem). Since we are typically interested in solving several problem instances, this puts an additional burden on the available time.
  \item[$\bullet$] Fourth, the search spaces in the context of combinatorial optimization tend to be several orders of magnitude larger than the number of game states for game playing. For instance, the number of game states for Go and chess are at most equal to $10^{171}$ \cite{Tromp:2016} and $10^{47}$ \cite{Chinchalkar:1996}, respectively, whereas the number of solutions for a problem instance with 1,000 binary decision variables is equal to $2^{1,000}>10^{300}$. However, the search spaces in the context of combinatorial optimization tend to have a certain structure that can be exploited to efficiently search them (e.g. using dominance and bounding rules). These differences motivate the enhancements that we propose, which are discussed next.
\end{description}

	\subsection{Domain reduction}
	\label{domainReductionSubsection}
	When the algorithm arrives in a node, several decision variables have already been instantiated. The edges to the children of the current node represent all the possible assignments of values to the current decision variable (i.e. the domain is enumerated). By using problem specific information, it is sometimes possible to prove that a certain value assignment to the current decision variable is \textit{dominated} by another value assignment, given the current partial solution. We make the concept of dominating more precise in the following definition:

\begin{definition}
  Assume (without loss of generality) that we are dealing with a minimization problem and that there are $n$ decision variables $a_1, a_2, \ldots, a_n$, which can each be instantiated by a value of their finite domains $d_1, d_2, \ldots, d_n$. Let $[a_1 \gets v_1, a_2 \gets v_2, \ldots, a_n \gets v_n]$ denote a solution where decision variable $a_i$ has been instantiated with value $v_i \in d_i$ ($\forall 1\leq i \leq n$) and let $f([v_1, v_2, \ldots, v_n])$ denote the objective function value associated with the solution $[a_1 \gets v_1, a_2 \gets v_2, \ldots, a_n \gets v_n]$ (equal to $\infty$ if the solution is infeasible). Consider a partial solution $[a_1 \gets v_1, a_2 \gets v_2, \ldots, a_k \gets v_k]$ where the first $k<n$ values have already been instantiated. We say that value assignment $a_{k+1} \gets v_{x_{k+1}}$ is dominated by value assignment $a_{k+1} \gets v_{y_{k+1}}$ (with $v_{x_{k+1}},  v_{y_{k+1}} \in d_{k+1}$), given the partial solution $[a_1 \gets v_1, a_2 \gets v_2, \ldots, a_k \gets v_k]$ if and only if 
$\\ \forall v_{x_{k+2}} \in d_{k+2}, v_{x_{k+3}} \in d_{k+3}, \ldots, v_{x_{n}} \in d_{n} : \exists v_{y_{k+2}} \in d_{k+2}, v_{y_{k+3}} \in d_{k+3}, \ldots, v_{y_{n}} \in d_{n} : \\
f([v_1, v_2, \ldots, v_k, v_{y_{k+1}}, v_{y_{k+2}}, \ldots, v_{y_{n}}]) \leq f([v_1, v_2, \ldots, v_k, v_{x_{k+1}}, v_{x_{k+2}}, \ldots, v_{x_{n}}])$.
\end{definition}

Hence, in the context of combinatorial optimization such children that lead to dominated solutions can be removed, despite the fact that they may represent feasible solutions.

	\subsection{Pruning subtrees by calculating bounds}
	When the algorithm reaches a node it has never visited before, a dual bound on the objective function is calculated to determine whether it is worthwhile to continue building a solution in the current iteration or not. The bound that is calculated is a lower bound for a minimization problem and an upper bound for a maximization problem, respectively. This bound is compared with the best solution value found so far and, in case it is impossible to find a solution that is better than this, the iteration is stopped. This bound is stored such that it can be used in further iterations without having to recalculate the bound. If the bound is worse than the current best objective function value, the current node is deleted as a child node of its parent node to avoid revisiting this node in further iterations (i.e. the subtree rooted at the current node is pruned). After deleting a child node, it is possible that a parent does not have any children left, which means in turn that there is no leaf node in the subtree rooted at the parent node that is better than the best solution found so far. In this case, we can also delete this parent node as a child node of its own parent node. This process can continue and hence deleting a node should be implemented in a recursive fashion such that the deletion of a node can give rise to the subsequent deletion of ancestors of this node.
	\subsection{Using a heuristic simulation policy}
	The default simulation policy for the most basic version of Monte Carlo tree search, as described in Section \ref{MCTSGamePlaying}, chooses children uniformly at random until a leaf node is reached. In the context of combinatorial optimization, however, often problem specific constructive heuristics are available to construct a solution. If one uses a heuristic for the solution completion policy of Monte Carlo tree search, the algorithm can be seen as an algorithm that attempts to learn to correct incorrect choices of the heuristic by selecting a different path in every iteration during the selection phase, corresponding with different value assignments to the decision variables. Furthermore, since the heuristic is executed starting from the root node of the tree in the first iteration of the algorithm, it is also guaranteed that the solution quality of the algorithm is at least as good as the solution quality of the heuristic.

\subsection{Selection Policy}
	\label{selectionSubsection}
	To decide which child to descend to in the selection phase, the algorithm keeps track of two values (for all nodes and edges, respectively): $numberVisits(v)$ and $averageObjectiveFunctionValue(v,w)$. Here, $numberVisits(v)$ denotes the number of times that node $v$ was visited and $averageObjectiveFunctionValue(v,w)$ denotes the average objective function value over all iterations in which the edge between parent node $v$ and child node $w$ belonged to the generated path. We constructed a similar expression as expression (\ref{argmaxExpression}) that is used in Monte Carlo tree search for game playing to select a child node. In this expression the first term is the average win ratio and is always between 0 and 1. We replace this term by a similar term (applicable to combinatorial optimization problems) that is always between 0 and 1 and that represents the quality of the obtained solutions. We define the value $normalizedScore(v,w)$ for all edges between parent node $v$ and child node $w$ as follows: let $v$ be a node, let $visitedChildren(v)$ denote the set of children of $v$ that have been visited at least once and let $unvisitedChildren(v)$ denote the set of children of $v$ that have not been visited before. We define $score(v,w)$ as the rank of node $w$ that we obtain by sorting the nodes in $visitedChildren(v)$ in increasing order of quality according to the value $averageObjectiveFunctionValue(v,w)$ (i.e. edges that seem more promising get a higher score). Now $normalizedScore(v,w)$ is defined as: 
\begin{equation}
\label{normalizedScoreExpression}
normalizedScore(v,w)=\frac{score(v,w)}{\sum_{w' \in visitedChildren(v)} score(v,w')}
\end{equation}
Hence, during the selection phase the algorithm will select the child for which the following expression is maximized \blue{(with ties broken in an arbitrary fashion)}:
\begin{equation}
\label{argmaxExpression2}
normalizedScore(parent,child)+\sqrt{\frac{2 \cdot \ln{(numberVisits(parent))}}{numberVisits(child)}}
\end{equation}
As mentioned in the previous section, the algorithm balances exploitation and exploration by maximizing this expression (the first term of this expression is high for promising value assignments, whereas the second term is high for infrequently used value assignments). We cope with the issue that this term is undefined for children that have not been visited before (because of division by 0) as follows: let $k_1=|unvisitedChildren(v)|$ and let $k_2=|visitedChildren(v)|$\blue{, where $|.|$ represents the size of a set}. To choose which child node to descend to, the algorithm selects with probability $\frac{k_1}{k_1+k_2}$ a child uniformly at random from the set of unvisited children and with probability $1-\frac{k_1}{k_1+k_2}$ the child for which expression (\ref{argmaxExpression2}) is maximized. 

A numerical example is given in Table \ref{numericalExampleTable} for a minimization problem in which a parent node has five children to choose from. The parent node has already been visited seven times and during these iterations, three different children were visited while two children have never been visited before. These three children have an average objective function value of 759.3, 753.0 and 751.3 (ordered in increasing order of quality). The second and the third column in this table ($numberVisits$ and $averageObjectiveFunctionValue$)  completely determine the values in all other columns.

\begin{table}[h!] \scriptsize\centering 
	\begin{threeparttable}
		\caption{Numerical example of calculating the selection probabilities for a minimization problem}
		\label{numericalExampleTable} 
		\begin{tabular}{ccccccccccccccccccccccccccccccccccccccccccccccccccc} \\
			\hline\noalign{\smallskip} 
			\multicolumn{1}{l}{} & \multicolumn{1}{l}{$numberVisits$} & \multicolumn{1}{l}{$averageObjectiveFunctionValue$} & \multicolumn{1}{l}{$score$} & \multicolumn{1}{l}{$normalizedScore$} & \multicolumn{1}{l}{expression (\ref{argmaxExpression2})} & \multicolumn{1}{l}{selection probability} \\ 
			\noalign{\smallskip}
			\hline
			\noalign{\smallskip} 
			\multicolumn{1}{c}{Child 1} & 3 & 751.3 & 3 & $\frac{3}{1+2+3}$ & 1.64 & 0 \% \\ 
			\multicolumn{1}{c}{Child 2} & 3 & 759.3 & 1 & $\frac{1}{1+2+3}$ & 1.31 & 0 \% \\ 
			\multicolumn{1}{c}{Child 3} & 1 & 753.0 & 2 & $\frac{2}{1+2+3}$ & 2.31 & 60 \%\\ 
			\multicolumn{1}{l}{Child 4} & 0 & - & - & - & - & 20 \% \\ 
			\multicolumn{1}{l}{Child 5} & 0 & - & - & - & - & 20 \% \\ 
			\noalign{\smallskip}
			\hline
			\noalign{\smallskip}
		\end{tabular}
	\end{threeparttable}
\end{table}

Special care must be taken when the values $averageObjectiveFunctionValue(v,w)$ are updated if an iteration produces an infeasible solution, because the objective function value of an infeasible solution is not meaningful. In this case, the algorithm will add a flag to the node that was added in the expansion phase, which prevents the algorithm from revisiting this node in any further iterations. Hence, the subtree rooted at this node is heuristically removed from the search space tree and the values $averageObjectiveFunctionValue(v,w)$ remain unchanged, because the parent node is regarded as having one less feasible child. This special case also illustrates the need for a well-performing heuristic simulation policy, because the algorithm punishes infeasible solutions by removing the node from which the heuristic simulation policy was started.

	\subsection{Directing the search by using a beam width}
	\label{beamWidthSubsection}
	To be able to learn for a given parent node which child node will lead to promising solutions, the child nodes should each be visited several times. However, the deeper the algorithm descends in the tree the more nodes there are at a given depth. For large search space trees it is not computationally feasible to visit all nodes several times, because if this were the case one could simply enumerate the whole search space. For this reason, we will shrink the search space to focus only on the most promising parts. This is done as follows: let $d$ be the number of decision variables (i.e. the depth of the search space tree) and $t$ be the number of seconds for which the algorithm is run. The algorithm will work in $d$ stages \blue{such that each stage takes} $\frac{t}{d}$ seconds. After stage $i$, the algorithm will only keep a beam of the $w$ most promising nodes at depth $i$ of the search space tree (according to the value $averageObjectiveFunctionValue$), where $w$ is an algorithm parameter that denotes the width of the beam. For the sake of computational efficiency, this should be implemented in a lazy way such that one should not explicitly delete all but the $w$ most promising nodes, but instead indicate for those $w$ nodes that they were not deleted and treat all other nodes as if they were deleted. Note that if there are at most $w$ nodes at a given depth, all nodes at this depth will be kept.

\subsection{Generalizability}
Although it is clear that the proposed algorithm could in principle be applied to any optimization problem where the decision variables all have finite domains, it might work better on certain problems than on others. In this subsection we will briefly highlight several desirable properties of a problem that make it well-suited for the proposed algorithm. First, the algorithm can be expected to perform better when the domains of the decision variables are not too big. This is the case, because in order to learn for a particular decision variable which value assignments tend to yield better solutions, all possible value assignments to that variable should be tried out at least once and preferably several times. Recall that for every edge $(v,w)$ the value $averageObjectiveFunctionValue(v,w)$ is stored to guide the algorithm in the selection phase and this value gets more accurate as more iterations are performed. Hence, the average branching factor of the tree should not be too high. However, the number of decision variables (i.e. the depth of the tree) is a somewhat less limiting factor, as will also become more clear in Section \ref{computationalResults}. Second, using stronger domain reduction rules will likely have a positive effect on the algorithm's performance and problems for which strong domain reduction rules are available are better suited for the algorithm. This is immediately linked to the first point, because strong domain reduction rules can eliminate more values from the domains of variables. Third, the tightness of the bounds used for pruning subtrees and the complexity of computing them both play an important role in the algorithm. In case tight bounds are available, large parts of the search space can be eliminated, allowing the algorithm to focus on more promising parts of the search space. The complexity of computing the bounds has an immediate effect on the number of iterations that the algorithm can perform in a given time. Usually it takes more time to compute stronger bounds, so it is important that a good trade-off between tightness and complexity is found.

	\section{Case study A: quay crane scheduling problem with non-crossing constraints}
	\label{caseStudy}
	The proposed algorithm was validated on two specific combinatorial optimization problems. In the first case study, we focus on the quay crane scheduling problem with non-crossing constraints. In the \blue{crane scheduling} literature, it is classified with the name $[1D||C_{Max}]$ according to the classification scheme proposed in \cite{Boysen:2017}. In this problem, we are given a set of $n$ bays $B=\{0, 1, \ldots, n-1\}$ on a container vessel and a set of $m$ quay cranes $K=\{0, 1, ..., m-1\}$. Both the bays and the quay cranes are placed from left to right on a one-dimensional line, where the lower numbered bays and quay cranes are more towards the left. An integer processing time $p_b$ is associated with every bay. The time it takes for the quay cranes to move parallel to the bays is assumed to be negligible in comparison with the processing times. The goal of the problem is to find a schedule that determines which quay crane has to process which bay at which time such that the makespan (the last completion time of a bay) is minimized. Furthermore, there are several constraints that need to be satisfied. The schedule has to be non-preemptive, which means that once a quay crane starts processing a bay, the quay crane keeps on processing this bay until it is finished. Hence, every bay is processed by exactly one quay crane. Because the quay cranes are located along a line from left to right, the quay cranes cannot cross each other. More specifically, if quay crane $k_1$ is processing bay $b_1$ at the same time as quay crane $k_2$ is processing bay $b_2$, with $k_1<k_2$, then it must hold that $b_1<b_2$. Finally, there are also constraints that indicate that at all times the quay cranes must leave enough space for the other quay cranes. An illustration of a toy problem instance and its solution is given in Figure \ref{GanntChartFigure}, which will be discussed in more detail in Subsection \ref{mathematicalModelSubsection}.

\subsection{Mathematical model}
\label{mathematicalModelSubsection}
To formalize the quay crane scheduling problem with non-crossing constraints, we use the mathematical model proposed by Santini et al. \cite{Santini:2014}. In this model, the problem variables are the number of bays $n$ in the set $B$, the number of quay cranes $m$ in the set $K$ and the processing times of the bays $p_b$ ($\forall b \in B$). The value $M$ denotes a sufficiently large constant. The decision variables are the binary variables $x_{bk}$, the binary variables $y_{bb'}$, the real variables $c_b$ and the real variable $c$. The meaning of these decision variables is as follows: $x_{bk}$ indicates whether bay $b$ will be processed by quay crane $k$, $y_{bb'}$ indicates whether the processing on bay $b$ finishes before the processing on bay $b'$ starts, $c_b$ denotes the completion time of bay $b$ and $c$ indicates the makespan. The model is given by:
\begin{mini!}|l|
	{}{& & \ \ c}{}{} \label{objectiveFunction}
	\addConstraint{c}{\geq c_b,}{\forall b \in B} \label{makeSpanConstraint}
	\addConstraint{c}{\geq \sum\limits_{b \in B}^{} x_{bk}p_b,}{\forall k \in K} \label{makeSpanConstraint2}
	\addConstraint{c_b}{\geq p_b,}{\forall b \in B} \label{completionTimeConstraint}
	\addConstraint{\sum\limits_{k \in K}^{} x_{bk}}{=1,}{\forall b \in B} \label{exactlyOneConstraint}
	\addConstraint{c_b}{\leq c_{b'}-p_{b'}+M(1-y_{bb'}),}{\forall b,b' \in B, b \neq b'} \label{defineYConstraint}
	\addConstraint{\sum\limits_{k \in K}^{} kx_{bk}-\sum\limits_{k \in K}^{} kx_{b'k}+1}{\leq M(y_{bb'}+y_{b'b}),}{\forall b,b' \in B, b < b'} \label{nonCrossingAndNotSimultaneousConstraint}
	\addConstraint{\sum\limits_{k \in K}^{} kx_{b'k}-\sum\limits_{k \in K}^{} kx_{bk}}{\leq b'-b+M(y_{bb'}+y_{b'b}),}{\forall b,b' \in B, b < b'} \label{idleCraneConstraint}
	\addConstraint{x_{bk}}{=0,}{\forall b \in B, k \in K, k>b} \label{pushLeftConstraint}
	\addConstraint{x_{bk}}{=0,}{\forall b \in B, k \in K, n-b<m-k} \label{pushRightConstraint}
	\addConstraint{x_{bk}}{\in \{0,1\},}{\forall b \in B, k \in K} \label{domainConstraint1}
	\addConstraint{y_{bb'}}{\in \{0,1\},}{\forall b,b' \in B, b \neq b'}  \label{domainConstraint2}
	\addConstraint{c_b}{\in \mathbb{R},}{\forall b \in B}  \label{domainConstraint3}
	\addConstraint{c}{\in \mathbb{R}}{}  \label{domainConstraint4}
\end{mini!}

	The meaning of these constraints is as follows. Constraints (\ref{makeSpanConstraint}) express that the makespan is at least as large as the completion time of all bays while constraints (\ref{makeSpanConstraint2}) express that the makespan is also at least as large as the sum of the processing times of all bays that are processed by the same quay crane. Constraints (\ref{completionTimeConstraint}) express that the completion time of a bay is at least as large as its processing time. The fact that every bay is processed by exactly one quay crane is enforced by constraints (\ref{exactlyOneConstraint}). Constraints (\ref{defineYConstraint}) define the decision variables $y_{bb'}$: they are equal to 0 if the processing on bay $b$ finishes later than the processing on bay $b'$ starts. Next, constraints (\ref{nonCrossingAndNotSimultaneousConstraint}) ensure both that a quay crane can only process one bay at a given time and that two quay cranes that are processing a different bay at the same time have not crossed each other. Constraints (\ref{idleCraneConstraint})-(\ref{pushRightConstraint}) all ensure that quay cranes have enough space at all times. More specifically, constraints (\ref{idleCraneConstraint}) enforce that two quay cranes that are working at the same time leave enough space (at least one bay) for all quay cranes that are in between them, while constraints (\ref{pushLeftConstraint}) and (\ref{pushRightConstraint}) enforce that quay cranes will not be pushed off the left and right side of the ship, respectively. Finally, constraints (\ref{domainConstraint1})-(\ref{domainConstraint4}) indicate the domain of the decision variables.
\subsection{Relaxation}
\label{Relaxation}
	The model above contains decision variables whose domain is not finite ($c_b$ and $c$) and hence it cannot \blue{be directly} used for the algorithm proposed in this article. However, the approach that we will take to solve this problem is to instead generate solutions for a relaxation of this problem (using a model whose decision variables are all discrete) and choose the best found solution for this relaxation that is also feasible for the original problem. The relaxation that we will consider, is the relaxation studied by Lim et al. \cite{Lim:2007}. This relaxation is obtained by taking the original model and removing the constraints that indicate that the quay cranes must leave enough space for other quay cranes (constraints (\ref{idleCraneConstraint})-(\ref{pushRightConstraint})). The decision variables of this relaxation are the same as the decision variables of the original problem, so doing this does not immediately help. However, Lim et al. have also shown that this relaxation is equivalent with another model in which all decision variables are discrete. For this equivalent model, the only decision variables are the $n$ decision variables $\sigma_b$ ($\forall b \in B$), which indicate which quay crane will process bay $b$. Hence, the domain of $\sigma_b$ is $K=\{0, 1, ..., m-1\}$. They have shown that if we know the values of these decision variables (i.e. if we know which quay cranes will process which bays), we can determine the schedule with minimal makespan by letting all quay cranes process their assigned bays from left to right, where each quay crane moves to its next assigned bay as soon as its path to the next assigned bay is free (no other quay crane is blocking its path). This schedule (and the resulting objective function value) can be calculated as follows: let $earliestTime[k][b]$ denote the earliest time that quay crane $k$ can move past bay $b$. Now all $earliestTime$ values can be computed with a time complexity of $O(n \cdot m)$ by dynamic programming by iteratively updating $earliestTime$ for subsequent bays $b$ (the outer loop of the algorithm) and quay cranes $k$ (the inner loop of the algorithm). The algorithm will first increase $earliestTime[k][b]$ by the processing time $p_b$ of bay $b$ in case bay $b$ is processed by quay crane $k$ (i.e. $\sigma_b = k$) and later it might be increased further in case  $earliestTime[k+1][b]>earliestTime[k][b]$, because every quay crane has to wait for the quay crane immediately to its right to avoid that quay cranes cross each other. This is shown in the pseudocode of Algorithm \ref{earliestTimeAlgorithm}:

\begin{algorithm}[h!]
  \FnC{\FRecursC{$m$, $n$, $\sigma$, $p$}}{
    \KwData{\vspace{-0.5mm}
		    \\$m$: the number of quay cranes
		    \vspace{-0.5mm}
		    \\$n$: the number of bays
		    \vspace{-0.5mm}
		    \\$\sigma$: an array representing the allocated quay crane for every bay
		    \vspace{-0.5mm}
		    \\$p$: an array representing the processing time for every bay}
		    \vspace{-0.5mm}
    \KwResult{\\The matrix $earliestTime$}
    \vspace{-0.5mm}
    \tcc{Start of code}
    \vspace{-0.5mm}
    {$earliestTime \gets emptyMatrix(m,n)$}\;
    \vspace{-0.5mm}
    \For{\forcondC}{
       \vspace{-0.5mm}
	\For{\forcondD}{
       \vspace{-0.5mm}
	\tcc{Initialization}
	\vspace{-0.5mm}
	\eIf{$b  \geq 1$}
		{
			\vspace{-0.5mm}
      			{$earliestTime[k][b] \gets earliestTime[k][b-1]$}\;
    		}
		{
			\vspace{-0.5mm}
			{$earliestTime[k][b] \gets 0$}\;
		}
	\vspace{-0.5mm}
	\tcc{Bay $b$ is processed by quay crane $k$}
	\vspace{-0.5mm}
	\uIf{$\sigma[b]==k$}
		{
			\vspace{-0.5mm}
      			{$earliestTime[k][b] \gets earliestTime[k][b]+p[b]$}\;
    		}
	\vspace{-0.5mm}
	\tcc{Quay crane $k$ has to wait for quay crane $k+1$}
	\vspace{-0.5mm}
	\uIf{$k+1<m$ and $earliestTime[k+1][b]>earliestTime[k][b]$}
		{
			\vspace{-0.5mm}
      			{$earliestTime[k][b] \gets earliestTime[k+1][b]$}\;
    		}
	}
    }
    \vspace{-0.5mm}
    {\Return $earliestTime$}\;
  }
  \caption{Computes $earliestTime$}
  \label{earliestTimeAlgorithm}
\end{algorithm}

Because every quay crane waits for the quay crane immediately to its right, it holds that $earliestTime[0][b] \geq earliestTime[1][b] \geq \ldots \geq earliestTime[m-1][b]$ ($\forall b \in B$). Hence, the makespan $c$ is equal to the earliest time that the first quay crane can move past the last bay ($earliestTime[0][n-1]$).

\begin{figure*}[h!]
	\centering
  \includegraphics[width=0.5\textwidth]{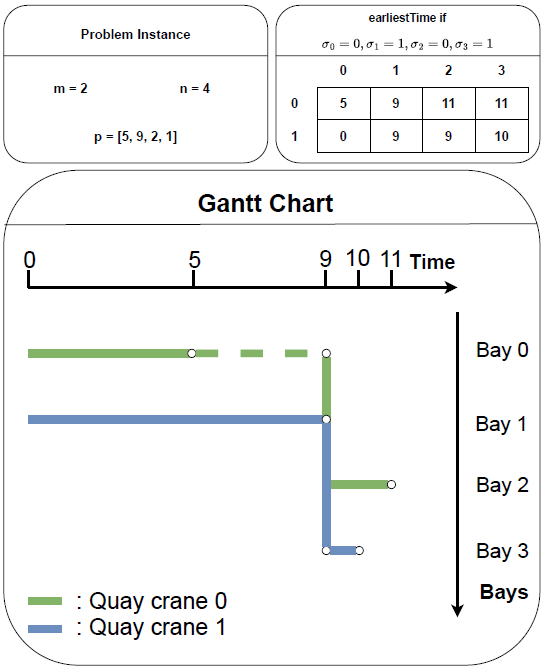}
\caption{A toy problem instance and an optimal solution with an objective function value of 11}
\label{GanntChartFigure}       
\end{figure*}

A toy problem instance and its optimal solution is given in Figure \ref{GanntChartFigure}. In this problem instance, there are $m=2$ quay cranes that have to process $n=4$ bays with a processing time of 5, 9, 2 and 1 time units (from left to right). A possible solution for this problem instance is for the first quay crane to process the first and the third bay and for the second quay crane to process the second and the fourth bay. The $earliestTime$ values are calculated according to the pseudocode of Algorithm \ref{earliestTimeAlgorithm}. For example, to calculate the $earliestTime$ values in the second column, the algorithm first sets $earliestTime[1][1]$ to $0+9=9$, because $earliestTime[1][0]=0$, $p[1]=9$ and $\sigma_1=1$.  The algorithm then sets $earliestTime[0][1]$ to 9, because $max(earliestTime[0][0], earliestTime[1][1])=9$. The corresponding schedule that is associated with this solution is depicted on the Gantt chart, where a dotted line represents that a quay crane is waiting and a full line represents that a quay crane is processing a bay. The makespan of this schedule is 11 (equal to $earliestTime[0][3]$ as desired), which is optimal. This solution is feasible for both the original problem (Subsection \ref{mathematicalModelSubsection}) and the relaxed version (Subsection \ref{Relaxation}).

\subsection{Enhancements used in our algorithm}
Finally, we will demonstrate for the relaxation above how we can enhance Monte Carlo tree search by reducing the domain of the decision variables, pruning subtrees by calculating bounds and using a heuristic solution completion policy.

\subsubsection{Domain reduction}
The generated solutions of the relaxation will be tested for feasibility for the original problem. Some of the constraints that were left out in the relaxation can however be directly satisfied by appropriately reducing the domains of the decision variables. More specifically, constraints (\ref{pushLeftConstraint}) and (\ref{pushRightConstraint}) are constraints that prohibit certain value assignments to the decision variables and can easily be satisfied by eliminating those values from the domains of the decision variables.

Another possibility for domain reduction makes use of Theorem \ref{domainReductionTheorem}, which was proven in \cite{Lim:2007}:
\begin{theorem}
\label{domainReductionTheorem}
Suppose that the values of the first $b$ decision variables $\sigma_0, \sigma_1, \ldots, \sigma_{b-1}$ are known and the algorithm currently has to choose the value for decision variable $\sigma_b$. If there exists a quay crane $k \in K$ such that $earliestTime[k][b-1]=earliestTime[k+1][b-1]$, then quay crane $k+1$ can be removed from the domain of $\sigma_b$ without changing the optimum. 
\end{theorem}
In order to reduce the domains by using this theorem, the $earliestTime$ values should be known before a complete solution is constructed, at the moment where the value for $\sigma_b$ is chosen. This can be done by updating the $earliestTime$ values on the fly, by executing the inner loop of Algorithm \ref{earliestTimeAlgorithm}, every time a value is assigned to a decision variable.
\subsubsection{Pruning subtrees by calculating bounds}
In the proposed algorithm, it is possible to prune subtrees by keeping track of the best solution found so far and comparing the associated objective function value with a lower bound on the objective function. In case the best found solution so far cannot be improved by further descending into the current subtree, it is pruned. Two lower bounds were proposed by Lim et al. \cite{Lim:2007} for the relaxation that was discussed earlier. Suppose that the values of the first $b$ decision variables $\sigma_0, \sigma_1, \ldots, \sigma_{b-1}$ are known and the algorithm currently has to choose the value for decision variable $\sigma_b$. Now, the first lower bound is given by:
\begin{equation}
\label{lowerBound1}
earliestTime[m-1][b-1]+\max(p_b, p_{b+1}, ..., p_{n-1})
\end{equation}
This bound can easily be proven by realising that every bay must be assigned to \blue{a} quay crane and the fact that $earliestTime[0][b] \geq earliestTime[1][b] \geq \ldots \geq earliestTime[m-1][b]$ $\forall b \in B$ (see earlier). Hence, assigning the remaining bay with the largest processing time to the earliest available quay crane is indeed a lower bound for the objective function value.

The second lower bound that was proposed, is given by:
\begin{equation}
\label{lowerBound2}
{\textstyle earliestTime[0][b-1]+\max \left( 0,\ceil*{\frac{\sum_{i=b}^{n-1} \left( p_i \right) - \sum_{i=1}^{m-1} \left(earliestTime[0][b-1]-earliestTime[i][b-1]\right)}{min(m,n-b)}} \right) }
\end{equation}
The fraction in this lower bound represents the following: first the sum of the processing times of all remaining bays is distributed over the quay cranes until the first quay crane does not have to wait anymore. After doing this, all the quay cranes have the same $earliestTime$ values, equal to $earliestTime[0][b-1]$. Now there are at most \blue{($n-b$)} bays left to process for $m$ quay cranes and the total remaining processing time is evenly spread over $min(m,n-b)$ quay cranes. Finally this result can be rounded up, because all processing times are integers and hence the real optimum has to be an integer as well. When spreading the processing times over the quay cranes, the constraint that every bay has to be assigned to exactly one quay crane is ignored and hence the obtained value is indeed a lower bound for the real optimal objective function value.

Both lower bounds can be calculated efficiently by doing an appropriate preprocessing step in $O(n)$ time, once before the start of the algorithm. The final lower bound that is used to compare against the current best solution is the maximum of both lower bounds. To calculate the first lower bound (expression (\ref{lowerBound1})), we have to be able to calculate $\max(p_b, p_{b+1}, ..., p_{n-1})$ efficiently without having to iterate over all the values inside the maximum. \blue{This can be achieved by calculating and storing the maximum of every suffix of $p_0, p_1, \ldots, p_{n-1}$ during a preprocessing step. These values can then be retrieved in $O(1)$ every time the lower bound is calculated.} Similarly, to calculate the second lower bound  (expression (\ref{lowerBound2})), we have to be able to calculate $\sum_{i=b}^{n-1} p_i$ and $\sum_{i=1}^{m-1} earliestTime[0][b-1]-earliestTime[i][b-1]$ efficiently. Here,  $\sum_{i=b}^{n-1} p_i$ can be calculated for every suffix in the preprocessing step, completely analogous to the approach that was used to calculate $\max(p_b, p_{b+1}, ..., p_{n-1})$. To calculate $\sum_{i=1}^{m-1} earliestTime[0][b-1]-earliestTime[i][b-1]$, however, the $earliestTime$ values depend on the choices that were made in the search space tree and hence cannot be calculated in the preprocessing step. Instead, a variable representing the value of this expression is stored and this value is updated on the fly when the $earliestTime$ values are changed.  
\subsubsection{Using a heuristic simulation policy}
When the algorithm arrives in a node which it has never visited before, the values for the remaining decision variables are chosen according to a heuristic. Suppose that the value of the first $b$ decision variables $\sigma_0, \sigma_1, \ldots, \sigma_{b-1}$ are already known, then the algorithm will subsequently choose the values for $\sigma_b, \sigma_{b+1}, \ldots, \sigma_{n-1}$ as follows. The algorithm consecutively tries to assign every value from the domain of $\sigma_b$ to $\sigma_b$ and updates the $earliestTime$ values and calculates the lower bound from the previous subsection. The final value that will be assigned to $\sigma_b$ is the value for which the calculated lower bound is as low as possible, which reflects the fact that we want the objective function value of the solution that is obtained in the end to be as low as possible. After doing this, the first $b+1$ decision variables are already instantiated and this process is repeated until all $n$ decision variables are instantiated.

\section{Case study B: 0-1 knapsack problem}
\label{caseStudyB}
In the second part of this case study, we focus on a classical problem: the 0-1 knapsack problem. In this problem, we are given a knapsack with capacity $c$ and $n$ items. With every item, we associate a positive integer profit $p_i$ and a positive integer weight $w_i$. The goal of the problem is to select a set of items to include in the knapsack such that the sum of the profits of these items is maximized and the sum of the weights of these items does not exceed the knapsack capacity. Hence, this problem can be written concisely as the following maximization problem:

\begin{maxi!}|l|
	{}{& & \ \ \sum_{i=1}^{n}p_i x_i}{}{} \label{objectiveFunction2}
	\addConstraint{\ \ \sum_{i=1}^{n}w_i x_i \leq c}{}{} \label{capacityConstraint}
	\addConstraint{\ \ x_{i} \in \{0,1\},}{}{\ \ \forall i \in \{1,2,\ldots,n\}} \label{domainConstraint5}
\end{maxi!}

Here, the problem variables are the capacity $c$, the number of items $n$, the profits of the items $p_i$ and the weights of the items $w_i$. The decision variables are the $n$ binary variables $x_i$. The meaning of these decision variables is as follows: $x_i$ is equal to 1 if and only if the $i$-th item is included in the knapsack. We will further assume that $x_i \leq c$ ($\forall i \in \{1,2,\ldots,n\}$) and $\sum_{i=1}^{n}w_i x_i > c$ to avoid trivial solutions. Finally, we also assume that the items are ordered according to the ratio between the profit and the weight such that:
\begin{equation}
\label{sortExpression}
\frac{p_i}{w_i} \geq \frac{p_{i+1}}{w_{i+1}}, \ \ \forall i \in \{1,2,\ldots,n-1\}
\end{equation}

This last assumption is introduced for convenience and will make the notation in the rest of the paper easier. In the following paragraphs, we will discuss for the case of the 0-1 knapsack problem which particular choices we have made for the enhancements that were proposed for the Monte Carlo tree search algorithm in the context of combinatorial optimization.

\subsection{Enhancements used in our algorithm}
\subsubsection{Domain reduction}
Suppose that the first $k<n$ decision variables $x_1, x_2, \ldots, x_k$ have already been instantiated. If $w_{k+1}+\sum_{i=1}^{k}w_i x_i> c$, we can eliminate the value 1 from the domain of $x_{k+1}$. This is the case, because if we assign 1 to $x_{k+1}$, the knapsack capacity is exceeded and the sum of the weights cannot decrease by adding zero or more items in the knapsack (recall that the weights are positive integers).

\subsubsection{Pruning subtrees by calculating bounds}
The 0-1 knapsack problem is a maximization problem. Hence, the bound that will be calculated for every node $v$ of the search space tree is an upper bound for the best objective function value amongst all leaf nodes in the subtree rooted at $v$. In case the current best objective function value is greater than or equal to this upper bound, the subtree rooted at $v$ can be pruned, because no solution in this subtree can be better than the current best solution.

It is clear that if the first $k<n$ decision variables $x_1,x_2,\ldots,x_k$ have already been instantiated, the algorithm should try to assign values to the remaining \blue{($n-k$)} decision variables such that the sum of their profits is maximized and the sum of their weights does not exceed $c-\sum_{i=1}^{k}w_i x_i$. Hence, assigning a value to the next decision variable $x_{k+1}$ gives rise to a new, slightly different knapsack problem instance in which one less item than before is available and the knapsack capacity is changed to $c-\sum_{i=1}^{k+1}w_i x_i$. Because of this reason, the upper bound in every node of the search space tree is an upper bound for a slightly different problem instance. The upper bound that the algorithm uses is due to Dantzig \cite{Dantzig:1957} and is obtained from solving the linear programming relaxation of the 0-1 knapsack problem. If constraint (\ref{domainConstraint5}) is replaced by $0 \leq x_i \leq 1, \forall i \in \{1,2,\ldots,n\}$ (with $x_i \in \mathbb{R}$), the resulting optimization problem can be solved exactly in a greedy manner by consecutively adding the largest possible fraction of the next item to the knapsack. More specifically, let $b$ be an index of an item such that $\sum_{i=1}^{b-1}w_i \leq c$ and $\sum_{i=1}^{b}w_i > c$. Item $b$ is called the break item (or sometimes also split item), because it is the first item that does not fit entirely in the knapsack. The optimal solution to the linear programming relaxation of the 0-1 knapsack problem is obtained by setting:

\begin{equation}
\begin{split}
x_i & = 1, \ \ \forall i \in \{1,2,\ldots,b-1\} \\
x_b & = \frac{c-\sum_{i=1}^{b-1}w_i}{w_b}  \\
x_i & = 0, \ \ \forall i \in \{b+1,b+2,\ldots,n\}
\end{split}
\end{equation}

The corresponding objective function value is given by:
\begin{equation}
p_b \cdot \frac{c-\sum_{i=1}^{b-1}w_i}{w_b}+\sum_{i=1}^{b-1}p_i
\end{equation}

As indicated earlier, this upper bound has to be calculated in all consecutive nodes of the search space tree that are on the path of a single iteration of the Monte Carlo tree search algorithm. Since the index of the break item $b$ can only increase as more decision variables are instantiated, the algorithm can avoid doing redundant computations by remembering the index of the break item for the previous node of a path and increasing this index for the current node if this is necessary. Since the index of the break item $b$ can only move at most $n$ times to the right and there are at most $n$ nodes on a path that is generated by the Monte Carlo tree search algorithm, all upper bounds for the path can be calculated efficiently with a total amortized time complexity of $O(n)$.

\subsubsection{Using a heuristic simulation policy}
When the algorithm arrives in a node that it has never visited before, the values for the remaining decision variables are chosen according to a heuristic. This heuristic considers the remaining items one by one and greedily adds an item to the knapsack if the item still fits in the knapsack. This heuristic is quite similar to the algorithm for the linear relaxation of the 0-1 knapsack problem which was discussed in the previous paragraph. However, the heuristic simulation policy will not add a fraction of the break item to the knapsack, but will instead leave the break item out of the knapsack and continue to add the remaining items one by one as long as there is still enough space. Because of the close resemblance between these two algorithms, the gap between the upper bound from the previous paragraph and the lower bound obtained by the heuristic simulation policy is usually very small. Later in Subsection \ref{computationResults01Knapsack}, it will become clear that this heuristic works remarkably well for many problem instances.

\section{Computational results}
\label{computationalResults}
	The proposed Monte Carlo tree search algorithm was extensively tested on two different case studies, requiring approximately 1,050 CPU-hours to run all the tests. In this section, we will consider five research questions: (1) What is the impact of varying the algorithm's parameters? (2) How does the performance of the algorithm compare with state-of-the-art algorithms? (3) Is the proposed algorithm able to improve the heuristic simulation policy? (4) What is the impact of the different components of the algorithm? (5) How do different selection policies affect the performance of the algorithm? We will answer all five questions for the first case study, but for the second case study we limit ourselves to the first four questions to limit the length of the paper. The results for the two case studies will be discussed in two different subsections.

To keep the computational burden manageable, the experiments were conducted on the ThinKing cluster of the Flemish Supercomputer Center (VSC), using powerful CPUs with a clock rate of 2.5 GHz and 10 GB RAM memory. The C++ code for our algorithms and the datasets that we generated are available online at\\ \textit{https://github.com/JorikJooken/MCTSQuayCraneSchedulingNonCrossingConstraints} and\\ \textit{https://github.com/JorikJooken/MCTS01Knapsack}. In the tables of this section, the numbers between brackets indicate the deviations from the best known dual bound (i.e. the dual bound refers to a lower bound for the quay crane scheduling problem with non-crossing constraints and an upper bound for the 0-1 knapsack problem) and the best found primal bounds (objective function values) for every experiment will be marked in bold. In the main section of this paper, we provide summaries of the experiments, whereas the tables in the appendix (Tables \ref{computationalResultsComparison}-\ref{detailedVersionComparisonKnapsack2}) contain the detailed results. For every table in the appendix, we also perform Wilcoxon signed-rank tests (non-parametric statistical hypothesis tests) in which we compare paired samples formed by the objective function value of the algorithm that we propose in the current paper (i.e. the reference algorithm, abbreviated as REF. ALG.) and the objective function value of another algorithm, which is different for every experiment. The null hypothesis of these tests states that the median of $X-Y$ is greater (less) than or equal to $0$ for the first case study (second case study), where $X$ and $Y$ are random variables that represent the objective function value of the reference algorithm and the other algorithm, respectively. In other words, for each case study small p-values in the tables indicate that the reference algorithm tends to produce a better objective function value more often than the other algorithm.

\subsection{Computational results: quay crane scheduling problem with non-crossing constraints}
\label{computationResultsQuayCraneScheduling}
\subsubsection{Data generation}
\label{dataGenerationQuayCraneScheduling}

The set of problem instances that we will use to answer these questions consists of two separate datasets (A and B). Dataset A was proposed by Lee and Chen \cite{Lee:2010} and later also used by Santini et al. \cite{Santini:2014}. Most of these problem instances represent a realistic situation with respect to the number of quay cranes and bays. To give the reader an idea of realistic numbers, we refer to Santini et al. \cite{Santini:2014} who report that there were 23 bays in one of the largest container vessels in 2,014 and Ng et al. \cite{Ng:2006} who report that two to seven quay cranes are deployed for a ship in a typical terminal. Apart from these realistic instances, this set also consists of a few problem instances for which the number of quay cranes and bays \blue{is} unrealistically large. The number of bays in problem instances from dataset A is between 16 and 100, while the number of quay cranes is between 4 and 10. The processing times of the bays are uniformly distributed between 30 and 100, leading to sums of processing times between 2,893 and 16,519.

Dataset B was generated by us in the same way as described in the previous paragraph, but consists of much bigger problem instances that represent unrealistic situations. The purpose of these unrealistically large problem instances is to demonstrate that heuristic algorithms are still able to produce good results in a reasonable time. The number of bays in these instances is between 200 and 3,000, the number of quay cranes is between 4 and 10 and the sums of the processing times are between 33,532 and 495,726.

\subsubsection{Experiments}
\label{experimentsQuayCraneScheduling}
In the first experiment, we have investigated the impact of varying the parameters of the Monte Carlo tree search algorithm (question (1)). This algorithm, which we will denote by $MCTS\_t\_w$, has two parameters that have to be chosen: the execution time $t$ (in seconds) and the beam width $w$. All combinations of parameters were tested, where $t \in \{10; 100\}$ and $w \in \{1; 10; 100\}$, yielding 6 different possibilities. The possible values for the execution time $t$ were chosen in such a way that they are comparable with the times that were reported in state-of-the-art algorithms (see the second experiment), while the values for the beam width $w$ were chosen to be reasonably small integers. Because of the stochasticity of the algorithm, every problem instance was solved 25 times for every parameter setting, each time using a different seed for the random number generator (a Mersenne Twister with a state size of 19,937 bits). The average objective function values over these 25 runs were recorded for every problem instance and these values were again averaged over the whole dataset. These values are summarized in Table \ref{computationalResultsVaryingParameters} for both datasets. Since the quay crane scheduling problem with non-crossing constraints is a minimization problem, lower values are better.

\begin{table}[h!] \scriptsize\centering 
	\begin{threeparttable}
		\caption{Average objective function values for varying parameters}
		\label{computationalResultsVaryingParameters} 
		\begin{tabular}{ccccccccccccccccccc} \\
			\hline\noalign{\smallskip} 
			\multicolumn{1}{l}{Dataset average} & \multicolumn{1}{l}{$MCTS\_10\_1$} & \multicolumn{1}{l}{$MCTS\_10\_10$} & \multicolumn{1}{l}{$MCTS\_10\_100$} & \multicolumn{1}{l}{$MCTS\_100\_1$} & \multicolumn{1}{l}{$MCTS\_100\_10$} & \multicolumn{1}{l}{$MCTS\_100\_100$} \\ 
			\noalign{\smallskip}
			\hline
			\noalign{\smallskip} 
			\multicolumn{1}{c}{Avg.\ A} & 847.9 & 847.4 & 846.9 & 845.7 & 845.5 & \textbf{845.3}\\
			\noalign{\smallskip}
			\hline
			\noalign{\smallskip}
			\multicolumn{1}{c}{Avg.\ B} & 38,195.2 & 38,192.8 & 38,193.0 & 38,180.0 & 38,180.5 & \textbf{38,178.5}\\
			\noalign{\smallskip}
			\hline
			\noalign{\smallskip}
		\end{tabular}
	\end{threeparttable}
\end{table}

	As can be seen from this table, the different parameter settings obtain very comparable results. The biggest relative differences between the different parameter settings are $0.31 \%$ and $0.04 \%$ for dataset A and B, respectively, where $MCTS\_10\_1$ had the worst performance and $MCTS\_100\_100$ had the best performance. For a fixed execution time $t$, the effect of varying the beam width $w$ is in general quite small and there is no single choice of beam width $w$ that is always at least as good as another choice for every single problem instance. For a fixed beam width $w$, the effect of changing the execution time $t$ is a little larger, although still quite small. On average, the relative improvement that could be gained by changing the execution time $t$ from 10 seconds to 100 seconds was $0.26 \%$, $0.22 \%$ and $0.19 \%$ (dataset A) and $0.04 \%$, $0.03 \%$ and $0.04 \%$ (dataset B) for a beam width of 1, 10 and 100, respectively. As we will see later in the paper (in Table \ref{summaryResults1}), the solutions produced by these six different algorithms are all nearly optimal (even using the worst parameter settings) and this explains why the observed differences are relatively small.

To answer the second question we have compared the performance of our algorithm (using the best obtained result from the previous experiment, where ties are broken in favour of less time in case of equal results) with the performance of the algorithms described by Lee and Chen \cite{Lee:2010}, Santini et al. \cite{Santini:2014} and Zhang et al. \cite{Zhang:2017}. Lee and Chen proposed a deterministic, heuristic algorithm, named \textit{enhanced best partition} (EBP), that consists of two phases. In the first phase, a more constrained version of the problem is solved in polynomial time by dynamic programming. In the second phase, this solution is iteratively improved by reassigning quay cranes to different bays. Santini et al. proposed the mathematical model that was introduced earlier in Subsection \ref{mathematicalModelSubsection} and used the mixed integer programming solver CPLEX to solve it. Zhang et al. proposed another deterministic, heuristic algorithm, named \textit{selected partition-based algorithm} (SPA), in which the bays are first partitioned and next consecutive areas of bays are merged according to several rules. The resulting algorithm achieves an approximation ratio of $2-\frac{2}{m+1}$, where $m$ denotes the number of quay cranes. This is the best approximation ratio that is known in literature.

We have contacted the corresponding authors of these three papers to inquire about the availability of the source code of their algorithms and the problem instances that were used. We have received the source code of Santini et al. \cite{Santini:2014} and the problem instances of Lee and Chen \cite{Lee:2010} and Santini et al. \cite{Santini:2014} (dataset A), but we were unable to obtain the source code of Lee and Chen \cite{Lee:2010} and Zhang et al. \cite{Zhang:2017}. Fortunately, these two missing algorithms were described in great detail without any unclear aspects in both papers and we have taken the effort to reimplement these algorithms. For the algorithm by Lee and Chen, we were able to verify that our reimplementation obtained exactly the same objective function values on the problem instances from dataset A as in the original paper \cite{Lee:2010}. The execution times are faster in our reimplementation, but this is very likely to be due to more modern hardware and a potentially faster programming language (MATLAB in the original paper versus C++ in our reimplementation). For the algorithm by Zhang et al. \cite{Zhang:2017}, there were unfortunately no common problem instances available for which we could test the correctness of our reimplementation. However, the optimality gaps that we obtained on our problem instances are comparable to the optimality gaps that were reported in \cite{Zhang:2017}, which gives us confidence that there were no \blue{significant errors} in our reimplementation of this algorithm.

The detailed results of these algorithms can be found in appendix A (Table \ref{computationalResultsComparison} for dataset A and Table \ref{computationalResultsComparison2} for dataset B) and are summarized in Table \ref{summaryResults1}. In these tables, the first column contains the name of the problem instance (or the dataset in case of Table \ref{summaryResults1}). In the remainder of this article, we will denote the problem instances by $n-m-s$, where $n$ represents the number of bays, $m$ represents the number of quay cranes and $s$ represents the sum of the processing times of all bays. These numbers are indicative of the size of the problem instance according to the mathematical model from Subsection \ref{mathematicalModelSubsection} (the higher these numbers, the larger the size). The second and the third column contain the lower bounds on the objective function values that were computed using expressions (\ref{lowerBound1})+(\ref{lowerBound2}) and CPLEX, respectively. Hence, the solutions that were computed using the different algorithms cannot have an objective function value that is below these lower bounds and equality would imply that the solution is optimal. Finally, the eight remaining columns contain the objective function value of the best found solution (with relative deviations from the best known lower bound indicated between brackets) and the execution time (with a maximum allotted time of 3,600 seconds) for all four algorithms. Table \ref{summaryResults1} contains the averages over the two datasets. In case not all algorithms were able to compute a feasible solution for every problem instance, we have made a distinction between the average over all problem instances (Avg.) and the average over all problem instances for which all algorithms produced a feasible solution (Avg.\ feasible). For dataset A, all algorithms were able to produce a feasible solution for every problem instance, but this was not the case for dataset B where CPLEX was not always able to find a feasible solution within 3,600 seconds.

\begin{table}[H] \scriptsize\centering 
	\setlength{\tabcolsep}{2.5pt}
	\begin{threeparttable}
		\caption{Summary of the comparison between the performances of  different algorithms for the quay crane scheduling problem with non-crossing constraints}
		\label{summaryResults1} 
		\begin{tabular}{cccccccccccccccccccccc} \\
			\hline\noalign{\smallskip} 
						\multicolumn{1}{c}{} & \multicolumn{1}{c}{} & \multicolumn{3}{c}{\thead{CPLEX \\(Santini et al. \cite{Santini:2014})}} & \multicolumn{2}{c}{\thead{MCTS \\(this paper)}} & \multicolumn{2}{c}{\thead{EBP \\(Lee and Chen \cite{Lee:2010})}} & \multicolumn{2}{c}{\thead{SPA \\(Zhang et al. \cite{Zhang:2017})}}\\
			\cmidrule(lr){3-5}\cmidrule(lr){6-7}\cmidrule(lr){8-9}\cmidrule(lr){10-11}
			\rot{Dataset average} & \rot{Lower bound (\ref{lowerBound1})+(\ref{lowerBound2})} & \rot{Lower bound} & \rot{Value} & \rot{Time} &  \rot{Value} & \rot{Time} & \rot{Value} & \rot{Time} & \rot{Value} & \rot{Time} \\ 
			\noalign{\smallskip}
			\hline
			\noalign{\smallskip} 
			\multicolumn{1}{c}{Avg.\ A} & 840.4 & 842.9 & 846.1 (0.4\%) & 1,937.7 s & \textbf{843.9 (0.1\%)} & 32.5 s &  883.1 (4.8\%) & $<$0.1 s & 1,327.6 (57.5\%) & $<$0.1 s\\ 	
			\noalign{\smallskip}
			\hline
			\noalign{\smallskip}
			\multicolumn{1}{c}{Avg.\ B} & 38,166.7 & - & - & - & \textbf{38,169.9 (0.0 \%)} & 51.3 s &  38,220.3 (0.1 \%) & 2.1 s & 64,239.2 (68.3 \%) & $<$0.1 s\\ 
			\multicolumn{1}{c}{\thead{Avg.\ B \\feasible}} & 7,137.0 & 7,137.0 & 33,487.8 (369.2 \%) & 3,600.0 s & \textbf{7,138.7 (0.0 \%)} & 47.5 s &  7,191.3 (0.8 \%) & $<$0.1 s& 11,941.3 (67.3 \%) & $<$0.1 s\\ 		
			\noalign{\smallskip}
			\hline
			\noalign{\smallskip}
		\end{tabular}
	\end{threeparttable}
\end{table}

	By inspecting Table \ref{summaryResults1} and Table \ref{computationalResultsComparison}, it becomes clear that for dataset A both EBP and SPA produce their answer very fast, while both MCTS and CPLEX take a longer time \blue{to} do so, but also obtain results with a higher quality. The results of SPA are quite different from the other three algorithms. The relative deviations from the lower bound range between $22.5 \%$ and $79.9 \%$ for SPA, while they are always below $15 \%$ for the other algorithms. This shows that although SPA obtains the best known approximation ratio ($2-\frac{2}{m+1}$), its practical performance is not so good. For EBP the relative deviations range from $1.2 \%$ to $13.5 \%$, while for both MCTS and CPLEX they are always below $1.5 \%$. Recall that both SPA and EBP are deterministic, heuristic algorithms such that executing them multiple times or increasing their computational budget will not affect the results that they produce. Hence, the best results for dataset A are produced by CPLEX (average relative deviation of $0.4 \%$) and MCTS (average relative deviation of $0.1 \%$). There were 9 problem instances for which the result that was produced by MCTS was strictly better than the result of CPLEX and 1 problem instance for which CPLEX was better than MCTS. For 8 out of these 9 cases, MCTS improved the previously best known solution in literature (for problem instance $23-4-3,544$, MCTS matches the previously best known solution that was also obtained in \cite{Santini:2014}).

The problem instances of dataset B (Table \ref{summaryResults1} and Table \ref{computationalResultsComparison2}) are much bigger in comparison with dataset A and in this case the results of CPLEX are no longer satisfactory, with relative deviations ranging between $145.6 \%$ and $807.7 \%$. For the 12 biggest problem instances, CPLEX was no longer able to even produce any solution or lower bound at all. For these bigger instances, the results obtained by MCTS were always strictly better than the results obtained by the other three algorithms. It should also be noted that the lower bounds that are used by MCTS (\blue{expressions} (\ref{lowerBound1})+(\ref{lowerBound2})) seem to be tighter for dataset B than for dataset A. For dataset A, MCTS produced a solution with an objective function value equal to the lower bound (i.e. an optimal solution) in 5 cases, while there were 13 such cases for dataset B. \blue{When} there are more bays it is easier to find a partition of the bays into contiguous areas such that the bay areas all roughly have the same processing time. Such a partition into contiguous areas gives rise to a feasible solution and since there are more partitions in case the number of bays rises, the quality of the best solution with this property also increases. This is illustrated by the fact that the results of EBP, which are based on finding the best partition into contiguous areas, also get better (average relative deviation of 4.8 \% for dataset A versus average relative deviation of 0.1 \% for dataset B). Because the lower bounds used by MCTS become tighter, its results also get slightly better in comparison with dataset A (average relative deviation of 0.1 \% for dataset A versus average relative deviation of 0.0 \% for dataset B). The results of SPA, however, get slightly worse (average relative deviation of 57.5 \% for dataset A versus average relative deviation of 68.3 \% for dataset B). Hence, it can be concluded that in general the best results are obtained by MCTS and it surpasses the state-of-the-art results for this case study. The statistical tests that we performed also support this. At a significance level of $\alpha = 5 \%$, we reject the null hypothesis of the Wilcoxon signed-rank test in favour of the alternative hypothesis (the p-values can be found in \blue{Tables} \ref{computationalResultsComparison} and \ref{computationalResultsComparison2}). Note that all p-values are very small.

	We answer the third question by providing evidence for the claim that MCTS is able to improve the heuristic that is used as its solution completion policy. Using the heuristic solution completion policy of MCTS as a standalone heuristic corresponds to stopping MCTS after its first iteration. This is the case because in the first iteration of the algorithm, the root node has never been visited before and hence the algorithm switches from using its selection policy to using its simulation policy. For this reason, the performance of MCTS is always at least as good as the performance of the heuristic solution completion policy. The averaged performance of the heuristic simulation policy for both datasets can be found in Table \ref{computationalResultsHeuristic}. We have also again added the best known lower bound and the results of $MCTS\_100\_100$ from the first experiment (because this parameter setting yielded the best average performance)  to this table to make it easier for the reader to follow the comparison. The relative deviations from the lower bound are indicated between brackets. The problem instances for which the heuristic solution completion policy found an infeasible solution have been left out for computing the averages. The best found value for every row is marked in bold.

\begin{table}[h!] \scriptsize\centering 
	\begin{threeparttable}
		\caption{Comparison between MCTS and its heuristic solution completion policy used as a standalone heuristic}
		\label{computationalResultsHeuristic} 
		\begin{tabular}{cccccccccccccccccccccccccccccccccccccc} \\
			\hline\noalign{\smallskip} 
			\multicolumn{1}{c}{Dataset average} & \multicolumn{1}{c}{Lower bound} & \multicolumn{1}{c}{$MCTS\_100\_100$} & \multicolumn{1}{c}{Time} & \multicolumn{1}{c}{Heuristic solution completion policy} & \multicolumn{1}{c}{Time} \\ 
			\noalign{\smallskip}
			\hline
			\noalign{\smallskip}
			\multicolumn{1}{c}{Avg.\ A feasible} & 828.7  & \textbf{830.8 (0.3 \%)} & 100.0 s & 922.9 (11.4 \%) & $<$0.1 s\\
			\noalign{\smallskip}
			\hline
			\noalign{\smallskip}
			\multicolumn{1}{c}{Avg.\ B feasible} & 42,805.3 &  \textbf{42,811.8 (0.0 \%)} & 100.0 s & 42,898.4 (0.2 \%) & $<$0.1 s\\
			\noalign{\smallskip}
			\hline
			\noalign{\smallskip}
		\end{tabular}
	\end{threeparttable}
\end{table}

	From this table, one can observe that the heuristic simulation policy can be executed very fast and the quality of the produced solution is moderate to good. In some cases, the solution produced by the heuristic simulation policy is infeasible (there were 4 out of 24 infeasible solutions for dataset A and 7 out of 24 for dataset B). Recall that the followed strategy of MCTS is to attempt to solve the original problem by solving a relaxation and discarding infeasible solutions. MCTS is able to learn to correct all of these solutions and is able to improve the heuristic on average by $10.0 \%$ for dataset A and by $0.2 \%$ for dataset B. Note that for dataset B, this improvement is quite small, because the performance of the heuristic solution completion policy is already quite good and there is less room for improvement. MCTS did not produce any infeasible solutions for both datasets. Hence, one can indeed conclude that MCTS can also be seen as a powerful method to improve the performance of a constructive heuristic algorithm.

	To answer the fourth question we implemented 16 different versions of MCTS, where every subset of the modifications that we proposed in Section \ref{MCTSCombinatorialOptimization} is omitted once (see Table \ref{versionComparison1}). The first version contains all of the modifications, whereas for the other versions at least one modification is omitted (so the first version corresponds to the version that was discussed in the previous paragraphs). The used parameters for every version are the same ($t=100$ and $w=100$ as in the previous experiment). The versions where a certain modification is not present work as follows. If a version does not follow the heuristic simulation policy, the simulation policy assigns domain values uniformly at random to the decision variables in the simulation phase of the algorithm. Not using the idea of directing the search by using a beam width corresponds to using an infinitely large beam width. Finally, not using dominance rules or pruning rules will result in no domain values or subtrees being pruned from the search space. The detailed results of this experiment can be found in appendix A (Tables \ref{detailedVersionComparisonQCSP1}-\ref{detailedVersionComparisonQCSP3} for dataset A and Tables \ref{detailedVersionComparisonQCSP4}-\ref{detailedVersionComparisonQCSP6} for dataset B) and are summarized in Table \ref{versionComparison1}. We can see that version 16 (where none of the modifications that we propose are present) performs consistently worse than the other versions. For dataset B, there were even 8 problem instances for which version 16 was not able to find any feasible solutions. We can also see that usually the versions which contain more modifications perform better, but this is not always the case. For example, for problem instance $16-5-2,893$ from dataset A we can see that version 10 has a worse objective function value than version 12. The versions which include the heuristic simulation policy also perform much better than the versions that exclude it. Finally, also note that the first version, which contains all of the modifications that we propose, always performs at least as \blue{well} as the other versions for both datasets for every problem instance, except for two problem instances ($21-5-3,033$ and $25-5-4,334$). Hence, the first version is clearly the best version for this case study. This is also supported by the statistical tests that we performed. For dataset A and B, there are respectively 13 and 15 (out of 15) MCTS variants for which we reject the null hypothesis of the Wilcoxon signed-rank test in favour of the alternative hypothesis (the p-values can be found in Tables \ref{detailedVersionComparisonQCSP1}-\ref{detailedVersionComparisonQCSP6}; they are often even below 0.01). The null hypothesis cannot be rejected at a significance level of $5 \%$ for version 5 (p-value of 0.063) and version 7 (p-value of 0.063) based on dataset A, although version 1 performs better on average and produces a provably optimal solution for all but two problem instances.


\begin{table}[H] \scriptsize\centering 
	\begin{threeparttable}
		\caption{Average objective function values for different versions of MCTS}
		\label{versionComparison1} 
		\begin{tabular}{cccccccccccccccccccccc} \\
			\hline\noalign{\smallskip} 
			\thead{MCTS \\Version} & \thead{Heuristic \\simulation policy} & Beam & Domain reduction & \thead{Pruning subtrees \\by calculating bounds} & Avg.\ A & Avg.\ B & \thead{Avg.\ B \\feasible}\\ 
			\noalign{\smallskip}
			\hline
			\noalign{\smallskip} 
			\multicolumn{1}{c}{1} & \cmark & \cmark & \cmark & \cmark & \textbf{843.9} & \textbf{38,169.9} & \textbf{13,385.5}\\ 
			\multicolumn{1}{c}{2} & \cmark & \cmark & \cmark & \xmark & 849.5 & 38,176.7 & 13,386.4\\ 
			\multicolumn{1}{c}{3} & \cmark & \cmark & \xmark & \cmark & 847.0 & 38,185.3 & 13,386.0\\ 
			\multicolumn{1}{c}{4} & \cmark & \cmark & \xmark & \xmark & 852.5 & 38,189.3 & 13,387.3\\ 
			\multicolumn{1}{c}{5} & \cmark & \xmark & \cmark & \cmark & 844.8 & 38,174.3 & 13,385.7\\ 
			\multicolumn{1}{c}{6} & \cmark & \xmark & \cmark & \xmark & 849.6 & 38,176.2 & 13,386.3\\ 
			\multicolumn{1}{c}{7} & \cmark & \xmark & \xmark & \cmark & 845.5 & 38,185.5 & 13,386.2\\ 
			\multicolumn{1}{c}{8} & \cmark & \xmark & \xmark & \xmark & 851.5 & 38,196.5 & 13,390.0\\ 
			\multicolumn{1}{c}{9} & \xmark & \cmark & \cmark & \cmark & 874.2 & - & 13,669.3\\
			\multicolumn{1}{c}{10} & \xmark & \cmark & \cmark & \xmark & 902.4 & - & 13,669.5\\ 
			\multicolumn{1}{c}{11} & \xmark & \cmark & \xmark & \cmark & 927.7 & - & 14,097.5\\ 
			\multicolumn{1}{c}{12} & \xmark & \cmark & \xmark & \xmark & 949.5 & - & 13,990.1\\ 
			\multicolumn{1}{c}{13} & \xmark & \xmark & \cmark & \cmark & 865.8 & - & 13,696.0\\ 
			\multicolumn{1}{c}{14} & \xmark & \xmark & \cmark & \xmark & 902.2 & - & 13,698.0\\ 
			\multicolumn{1}{c}{15} & \xmark & \xmark & \xmark & \cmark & 912.9 & - & 13,954.8\\ 
			\multicolumn{1}{c}{16} & \xmark & \xmark & \xmark & \xmark & 937.0 & - & 14,105.3\\  	
			\noalign{\smallskip}
			\hline
			\noalign{\smallskip}
		\end{tabular}
	\end{threeparttable}
\end{table}

Finally to answer the fifth research question, we compared the selection policy that we propose in the current paper with two alternative ones. For the first alternative, we replaced expression (\ref{argmaxExpression2}) used to select a child in the selection phase with the following expression (which is more close to expression (\ref{argmaxExpression}) from Monte Carlo tree search in the context of game playing):

\begin{equation}
\label{argmaxExpression3}
averageObjectiveFunctionValue(parent,child)+\sqrt{\frac{2 \cdot \ln{(numberVisits(parent))}}{numberVisits(child)}}
\end{equation}

For the second alternative, we implemented\footnote{\blue{We used the same parameters as those used for the inventory control problem in \cite{Li:2021}, namely $n_0=2$, $\sigma_0^2=100$ and $\alpha_{N(\mathbf{x})} = 1-\frac{1}{5N(\mathbf{x})}$.}} the OCBA-MCTS selection policy proposed by Li et al. \citep{Li:2021}. This is based on the Optimal Computing Budget Allocation (OCBA) framework \citep{Chen:2000} and has nice theoretical properties. More specifically, Li et al. \citep{Li:2021} prove that the algorithm optimally allocates a limited computing budget such that a lower bound on the probability of correctly selecting the best action at each node is maximized. Note that this goal is different from the MCTS variant used in the current paper (which attempts to maximize the cumulative rewards).

The results of the three algorithms considered in this experiment are summarized in Table \ref{differentSelectionPolicies1} (see Tables \ref{differentSelectionPolicies2} and \ref{differentSelectionPolicies3} for the detailed results). For each algorithm, we again use the best found parameter setting from the first experiment ($w=100$ and $t=100$). As we can see, the three algorithms have a relatively comparable performance in the sense that all algorithms tend to produce nearly optimal solutions (the average deviation from the lower bound is at most $0.3 \%$ for all algorithms). The best average objective function values for both data sets are given by the algorithm which uses expression (\ref{argmaxExpression2}) from the current paper, the algorithm which uses expression (\ref{argmaxExpression3}) and OCBA-MCTS (in that order). The algorithm from the current paper produces a solution that is always at least as good as the other two algorithms on all problem instances, except for problem instance $25-5-4,334$. The Wilcoxon signed-rank test also confirms for both datasets that at a significance level of $\alpha = 5 \%$ the null hypothesis should be rejected in favour of the alternative hypothesis (the p-values can be found in Tables \ref{differentSelectionPolicies2} and \ref{differentSelectionPolicies3}).
\begin{table}[H] \scriptsize\centering
\setlength{\tabcolsep}{2.5pt} 
	\begin{threeparttable}
		\caption{Comparsion between different selection policies of MCTS.}
		\label{differentSelectionPolicies1} 
		\begin{tabular}{cccccccccccccccccccccc} \\
			\hline\noalign{\smallskip} 
			Dataset average & Lower bound & \thead{MCTS\\(selection policy uses (\ref{argmaxExpression2}))} & \thead{MCTS\\(selection policy uses (\ref{argmaxExpression3}))} & \thead{OCBA-MCTS\\(Li et al. \cite{Li:2021})}\\ 
			\noalign{\smallskip}
			\hline
			\noalign{\smallskip} 
			\multicolumn{1}{c}{Avg.\ A} & 842.9 & \textbf{843.9  (0.1\%)} & 844.7  (0.2\%)& 845.1 (0.3\%)\\
			\multicolumn{1}{c}{Avg.\ B} &38,166.7 & \textbf{38,169.9  (0.0\%)} & 38,178.3  (0.0\%)& 38,178.5 (0.0\%)\\
			\noalign{\smallskip}
			\hline
			\noalign{\smallskip}
		\end{tabular}
	\end{threeparttable}
\end{table}

\subsection{Computational results: 0-1 knapsack problem}
\label{computationResults01Knapsack}
\subsubsection{Data generation}
For the 0-1 knapsack problem, we have also generated two datasets of problem instances (dataset A and dataset B). It is well known that the 0-1 knapsack problem can be exactly solved with a worst-case time complexity of $O(n \cdot c)$ (see e.g. \cite{Pisinger:1997}), where $n$ denotes the number of items in the knapsack and $c$ denotes the knapsack capacity. Therefore, most of the problem instances were generated such that $n \cdot c$ is large.

The problem instances from dataset A are strongly correlated spanner instances, which were introduced by Pisinger in \cite{Pisinger:2005}. This article has introduced several classes of problem instances that were empirically shown to be hard to solve. Amongst all these classes, the strongly correlated spanner instances were the hardest class of problem instances. The weights and the profits of the $n$ items in a strongly correlated spanner instance are all multiples of the weights and profits of a small set of items (the spanner set). Pisinger has described in \cite{Pisinger:2005} that the problem instances become harder for smaller spanner sets, leading us to use 2 items for the spanner set. The weights $w_i$ of the items in the spanner set are generated independently from each other, uniformly at random between $1$ and $10^8$. The corresponding profits $p_i$ of the items in the spanner set are strongly correlated with the weights and are generated by setting $p_i = w_i+10^7$. The remaining \blue{($n-2$)} items are generated by selecting a random item $j$ from the spanner set and generating an integer multiplier $k$ uniformly at random between $1$ and $10$ for every item. The profit and weight of the $i$-th item are then obtained by setting $p_i = k \cdot p_j$ and $w_i = k \cdot w_j$\blue{, respectively}. Finally, the knapsack capacity is chosen to be a certain fraction $f$ of the sum of the weights of all the items. In the instances from dataset A, we have chosen $f \in \{0.25; 0.50; 0.75\}$ and $n \in \{50; 200; 500; 2,000; 5,000; 20,000; 50,000\}$.

The state-of-the-art Combo algorithm for the 0-1 knapsack problem \cite{Martello:1999} is still able to exactly solve the largest problem instances from dataset A in a reasonable time, despite the fact that these instances are the hardest problem instances described in \cite{Pisinger:2005}. For dataset B on the contrary, we have been able to create problem instances which consist of relatively few items in comparison with dataset A, but the time needed by the Combo algorithm to solve these problem instances is considerably higher. These problem instances are so-called noisy multi-group exponential problem instances and these instances have been shown to be difficult to solve for exact algorithms in \cite{Jooken:2022}. In the current paper we show that our heuristic algorithm is able to quickly produce nearly optimal solutions. An instance is generated as follows: the knapsack capacity $c$ is equal to $10^{10}$. There are $n$ items, which can be divided into 10 different groups. The first 9 groups (groups $1, 2, \ldots, 9$) consist of items with exponentially decreasing weights and profits. These 9 groups consist of around $\frac{2}{3} \cdot n$ items, where every group is equally large. All items in a certain group $i$ ($i=1,\ldots,9$) are generated independently from each other, by setting the profit $p_j$ of item $j$ as $p_j = (\frac{1}{2^i}+10^{-4}) \cdot c+r_1$ and setting the weight as $w_j = (\frac{1}{2^i}+10^{-4}) \cdot c+r_2$ (note that $p_j$ and $w_j$ are integers because of the choice of $c$). Here, $r_1$ and $r_2$ are small integers which are chosen uniformly at random between 1 and 300 for each item. Note that by defining the items precisely like this, the profit-weight ratios of the items in the first 9 groups are all very close to each other. Another consequence of the definition is that the optimal solution usually has to combine items from several different groups, because including $2^i$ items from group $i$ would slightly exceed the knapsack capacity, whereas including $2^i-1$ items from group $i$ would leave a large part of the knapsack unfilled. Finally the tenth group consists of the remaining $\frac{1}{3} \cdot n$ items and these items have very small weights and profits with a very diverse range of profit-weight ratios. These items are generated independently from each other by setting the profit $p_j$ of item $j$ as $p_j = r_1$ and setting the weight as $w_j = r_2$, where $r_1$ and $r_2$ are again small integers which are chosen uniformly at random between 1 and 300 for each item. Hence, the items in the tenth group introduce more \blue{variability} in the profit-weight ratios and they are useful to fill a small part of the knapsack. The problem instances from dataset B were generated by choosing $n \in \{100; 125; 150; \ldots; 700\}$.

\label{dataGeneration01Knapsack}
\subsubsection{Experiments}
\label{experiments01Knapsack}

To answer the first research question from Section \ref{computationalResults}, we investigated to which extent the results are affected by different parameter settings of the Monte Carlo tree search algorithm. As in the previous case study, we tested 6 different parameter settings that were obtained by choosing the beam width $w \in \{1; 10; 100\}$ and the execution time $t \in \{10; 100\}$. For every parameter setting, we solved every instance 25 times independently from each other, using a different random seed in every run. For both datasets, we computed the average objective function values over these 25 runs for every problem instance, and next these values were averaged over the whole dataset. The results of this experiment can be found in Table \ref{computationalResultsVaryingParameters2}. Recall that the 0-1 knapsack problem is a maximization problem, such that higher objective function values are better. The best found average for every dataset is marked in bold.

\begin{table}[h!] \scriptsize\centering 
	\begin{threeparttable}
		\caption{Objective function values for varying parameters}
		\label{computationalResultsVaryingParameters2} 
		\begin{tabular}{ccccccccccccccccccc} \\
			\hline\noalign{\smallskip} 
			\multicolumn{1}{l}{Dataset average} & \multicolumn{1}{l}{$MCTS\_10\_1$} & \multicolumn{1}{l}{$MCTS\_10\_10$} & \multicolumn{1}{l}{$MCTS\_10\_100$} & \multicolumn{1}{l}{$MCTS\_100\_1$} & \multicolumn{1}{l}{$MCTS\_100\_10$} & \multicolumn{1}{l}{$MCTS\_100\_100$} \\ 
			\noalign{\smallskip}
			\hline
			\noalign{\smallskip} 
			\multicolumn{1}{c}{Avg.\ A} & \textbf{334,940,368,920.9} & \textbf{334,940,368,920.9} & \textbf{334,940,368,920.9} & \textbf{334,940,368,920.9} & \textbf{334,940,368,920.9} & \textbf{334,940,368,920.9}\\
			\noalign{\smallskip}
			\hline
			\noalign{\smallskip}
			\multicolumn{1}{c}{Avg.\ B} & 9,999,791,698.2 & 9,999,846,037.4 & 9,999,721,277.9 & 9,999,875,981.1 & \textbf{9,999,892,314.5} & 9,999,829,188.7\\
			\noalign{\smallskip}
			\hline
			\noalign{\smallskip}
		\end{tabular}
	\end{threeparttable}
\end{table}

Interestingly, we see that changing the parameter settings has no effect at all for dataset A. The obtained objective function values are all exactly the same for the 6 different parameter settings and the 25 different runs, given a fixed problem instance of dataset A. For dataset B, different parameter settings yield different results, but only very slightly. For a fixed execution time $t$, the results are ordered in increasing order of quality by choosing the beam width $w$ as 100, 1 and 10 (in this order). Changing the execution time $t$ from 10 seconds to 100 seconds for a fixed beam width $w$ improves the results, albeit only slightly (the biggest relative difference with respect to the objective function value is equal to $1.1 \times 10^{-3}$ \%). These results reveal that the best parameter settings for both datasets are obtained by choosing $t=100$ and $w=10$, although the differences are very small.

We answer the second question by comparing the performance of our algorithm (using the best obtained result from the previous experiment, where ties are broken in favour of less time in case of equal results) with the performance of the famous Combo algorithm \cite{Martello:1999} (source code\footnote{We changed the value of the variable \textit{MAXSTATES} from $1.5 \times 10^6$ to $4.5 \times 10^8$ \blue{to avoid the program crashing when the number of states in the dynamic programming algorithm exceeds this variable.}} available online at: http://hjemmesider.diku.dk/{\texttildelow}pisinger/codes.html). Although this algorithm was invented more than twenty years ago, it still represents the current state-of-the-art (see e.g. \cite{Buther:2012}, \cite{Monaci:2013}, \cite{Pisinger:2017} and \cite{Huerta:2020} for relatively recent articles that support this claim). This is not surprising, given that the authors of Combo have been conducting research about the 0-1 knapsack problem for at least two decades prior to the publication of the article in which Combo was described \cite{Martello:1999}. The stellar performance of this algorithm is reconfirmed in our experiments. From these experiments, it becomes clear that many very large knapsack problem instances can be solved in only a few seconds, despite Combo being an exact algorithm. Because of the very limited room for improvement, there do not exist heuristic algorithms which are able to outperform Combo in this case. However, as will be indicated later, even the simple heuristic solution completion policy that we used in our algorithm produces objective function values that are extremely close to the optimal values. With this knowledge, the second research question is less interesting for the 0-1 knapsack problem (Combo has been the undisputed best algorithm for more than two decades), but we want to answer it anyways because we also addressed this question for the first case study. Nevertheless, we are also able to show that certain problem instances with specific characteristics exist (dataset B) for which the large execution times might make it impractical to use Combo.

The detailed results of Combo and MCTS can be found in appendix B (Table \ref{ComboMCTSComparisonA} for dataset A and Table \ref{ComboMCTSComparisonB} for dataset B) and are summarized in Table \ref{summaryResults2}. The first column contains the name of the problem instance (or name of the dataset in case of Table \ref{summaryResults2}). The instances will be denoted as strCorrSpan-$n$-$c$ and exp-$n$-$c$ for dataset A and B, respectively, where $n$ denotes the number of items in the knapsack and $c$ denotes the knapsack capacity. The four other columns contain the found objective function value and the time needed to obtain this result by both algorithms. Table \ref{summaryResults2} contains the averages over both datasets. The Combo algorithm computes the exact optimum while MCTS is a heuristic, and thus the obtained result by MCTS is a lower bound for this optimum. The relative deviation from the optimum is indicated between brackets.

\begin{table}[h!] \scriptsize\centering 
	\begin{threeparttable}
		\caption{Summary of the comparison between the performances of Combo and MCTS for the 0-1 knapsack problem}
		\label{summaryResults2} 
		\begin{tabular}{cccccccccccccccccccccc} \\
			\hline\noalign{\smallskip} 
			\multicolumn{1}{c}{} & \multicolumn{2}{c}{Combo} & \multicolumn{2}{c}{MCTS}\\
			\cmidrule(lr){2-3}\cmidrule(lr){4-5}
			\noalign{\smallskip} 
			Dataset & Value & Time & Value & Time \\ 
			\noalign{\smallskip}
			\hline
			\noalign{\smallskip} 
			\multicolumn{1}{c}{Avg.\ A} & \textbf{334,942,167,924.2} & 2.8 s & 334,940,368,920.9 ($5.3 \times 10^{-4}$ \%) & 10.0 s \\ 	
			\noalign{\smallskip}
			\hline
			\noalign{\smallskip}
			\multicolumn{1}{c}{Avg.\ B} & \textbf{9,999,968,161.2} & 866.9 s & 9,999,920,718.0 ($4.7 \times 10^{-4}$ \%) & 67.6 s \\ 	
			\noalign{\smallskip}
			\hline
			\noalign{\smallskip}
		\end{tabular}
	\end{threeparttable}
\end{table}

	If we take a closer look at Table \ref{ComboMCTSComparisonA}, we see that Combo is able to compute the optimal answer in less than 10 seconds for almost all problem instances of dataset A, except for the three largest problem instances where there are 50,000 items in the knapsack. For two of these three instances, it obtains a better objective function value than MCTS and requires more or less the same time to do so. However, it should also be noted that the objective function value produced by MCTS is always very close to the optimal answer: amongst all problem instances of dataset A the largest relative deviation from the optimum is equal to $0.029$ \% and the average relative deviation is equal to $5.3 \times 10^{-4}$ \%. For 14 problem instances, MCTS was able to produce an objective function value that was equal to the optimal one. Thus we can conclude that for dataset A both algorithms produce very similar solutions, but Combo has a slight advantage.

However, the conclusion for dataset B is somewhat different. These results can be found in Table \ref{ComboMCTSComparisonB}. Despite the fact that there are only several hundreds of items, the execution time needed by Combo to produce the optimal answer increases rapidly. For example, it took Combo more than 2,500 seconds to solve exp-700-10,000,000,000 from dataset B, while it took less than 0.1 second to solve strCorrSpan-2,000-17,052,969,836 from dataset A (note that the number of items $n$ and the knapsack capacity $c$ of both problem instances have a comparable order of magnitude).  For the problem instances of dataset B, MCTS is still able to produce objective function values that are very close to the optimal ones, while only using at most 100 seconds to do so. Hence, for these instances it might be advisable to use a heuristic like MCTS that sacrifices the optimality guarantee for a decrease in execution time. On average, MCTS used 67.6 seconds of execution time, while Combo used 866.9 seconds on average. The cost of this decrease in execution time is not so big: the average optimality gap attained by MCTS is equal to $4.7 \times 10^{-4}$ \%.

\begin{table}[h!] \scriptsize\centering 
	\begin{threeparttable}
		\caption{Comparison between MCTS and its heuristic solution completion policy used as a standalone heuristic}
		\label{computationalResultsHeuristic2} 
		\begin{tabular}{cccccccccccccccccccccccccccccccccccccc} \\
			\hline\noalign{\smallskip} 
			\multicolumn{1}{c}{Dataset average} & \multicolumn{1}{c}{Optimum} & \multicolumn{1}{c}{$MCTS\_100\_10$} & \multicolumn{1}{c}{Time} & \multicolumn{1}{c}{Heuristic solution completion policy} & \multicolumn{1}{c}{Time} \\ 
			\noalign{\smallskip}
			\hline
			\noalign{\smallskip}
			\multicolumn{1}{c}{Avg.\ A} & 334,942,167,924.2  & \textbf{334,940,368,920.9 ($\mathbf{5.3 \times 10^{-4}}$ \%)} & 100.0 s & 334,940,285,925.1 ($5.6 \times 10^{-4}$ \%)& $<$0.1 s\\
			\noalign{\smallskip}
			\hline
			\noalign{\smallskip}
			\multicolumn{1}{c}{Avg.\ B} & 9,999,968,161.2 & \textbf{9,999,920,718.0 ($\mathbf{4.7 \times 10^{-4}}$ \%)} & 100.0 s & 9,992,194,843.4 ($7.7 \times 10^{-2}$ \%) & $<$0.1 s\\
			\noalign{\smallskip}
			\hline
			\noalign{\smallskip}
		\end{tabular}
	\end{threeparttable}
\end{table}

To answer the third question, we compare $MCTS\_100\_10$ (the best parameter setting in the first experiment) with the heuristic that is used as its solution completion policy. Recall that the result of MCTS will always be at least as good as the result of its solution completion policy. The average performance of both algorithms can be found in Table \ref{computationalResultsHeuristic2}. We also repeated the average optimal objective function values and the relative deviations between brackets for the reader's convenience. The heuristic solution completion policy can be executed very fast (in less than 0.1 second), because it corresponds to the first iteration of MCTS. From this table we can also see that the objective function values obtained by the heuristic solution completion policy are very close to the optimal ones. For both datasets, MCTS is able to improve these values. For dataset A, this improvement is not so big (from $5.6 \times 10^{-4}$ \% to $5.3 \times 10^{-4}$ \%), because the initial optimality gap is already very small. For dataset B, the relative improvement is bigger. The optimality gap is reduced from $7.7 \times 10^{-2}$ \% to $4.7 \times 10^{-4}$ \%, which corresponds to a decrease with a factor of approximately 163.8. We conclude that MCTS is indeed able to learn to correct the solutions produced by its heuristic simulation policy, despite the fact that this policy already achieves small optimality gaps.

	Finally, to answer the fourth question we implemented five different versions of MCTS. The modifications that we proposed in Section \ref{MCTSCombinatorialOptimization} are introduced one by one in these five versions. The used parameters for every version are the same ($t=100$ and $w=10$ as in the previous experiment). The detailed results of this experiment can be found in appendix B (Table \ref{detailedVersionComparisonKnapsack1} for dataset A and Table \ref{detailedVersionComparisonKnapsack2} for dataset B) and are summarized in Table \ref{versionComparison2}. From these tables, it becomes clear that the fifth version of MCTS (where none of the modifications that we propose are present) is not suitable at all for this problem and it performs much worse than the four other versions. It can often not improve the initial solution in which none of the items are selected. This was the case for 7 problem instances of dataset A and all problem instances of dataset B. The four other versions were always able to produce nearly optimal solutions. For dataset A, there was only one problem instance ($strCorrSpan-2,000-28,036,577,373$) for which the solution produced by the four other versions was different. Hence, for dataset A these four other versions were equally good (apart from this one problem instance where version 1 and version 4 produced an equally good solution that was better than version 2 and version 3). For dataset B, however, the solutions produced by the four other versions were often slightly different. For this dataset, version 1 produced the best solutions on average and there were 14 problem instances for which it was strictly better than all other versions. However, this dataset also shows that version 1 was not always at least as good as the other versions (as was previously the case). The first version tends to produce a better solution most of the time. This was also confirmed by the Wilcoxon signed-rank test: at a significance level of $\alpha=5 \%$, the null hypothesis should be rejected in favour of the alternative hypothesis (the p-values can be found in Table \ref{detailedVersionComparisonKnapsack2}).

\begin{table}[H] \scriptsize\centering 
	\begin{threeparttable}
		\caption{Average objective function values for different versions of MCTS}
		\label{versionComparison2} 
		\begin{tabular}{cccccccccccccccccccccc} \\
			\hline\noalign{\smallskip} 
			Dataset average & MCTS Version 1 & MCTS Version 2 & MCTS Version 3 & MCTS Version 4 & MCTS Version 5\\ 
			\noalign{\smallskip}
			\hline
			\noalign{\smallskip} 
			\multicolumn{1}{c}{Heuristic simulation policy}& \cmark& \cmark& \cmark& \cmark& \xmark\\
			\multicolumn{1}{c}{Beam}& \cmark& \cmark& \cmark& \xmark& \xmark\\
			\multicolumn{1}{c}{Domain reduction}& \cmark& \cmark& \xmark& \xmark& \xmark\\
			\multicolumn{1}{c}{Pruning subtrees by calculating bounds}& \cmark\ & \xmark& \xmark& \xmark& \xmark\\
			\noalign{\smallskip}
			\hline
			\noalign{\smallskip}
			\multicolumn{1}{c}{Avg.\ A}& \textbf{334,940,368,920.9}& 334,940,326,098.6& 334,940,326,098.6& \textbf{334,940,368,920.9}& 187,232,935,624.5\\
			\noalign{\smallskip}
			\hline
			\noalign{\smallskip} 	
			\multicolumn{1}{c}{Avg.\ B}& \textbf{9,999,920,718.1}& 9,999,914,190.1& 9,999,914,267.6& 9,999,697,742.2& 0.0\\
			\noalign{\smallskip}
			\hline
			\noalign{\smallskip}
		\end{tabular}
	\end{threeparttable}
\end{table}

\section{Conclusions and further work}
\label{conclusions}
	In this article, we have proposed a heuristic algorithm to explore search space trees that is based on Monte Carlo tree search. By leveraging the combinatorial structure of the problem, the algorithm was enhanced in several ways. These enhancements were demonstrated on two case studies: the quay crane scheduling problem with non-crossing constraints and the 0-1 knapsack problem. The computational results for these problems have shown that the proposed algorithm is able to compete with state-of-the-art algorithms for both problems and eight new best solutions were found for the set of problem instances proposed by Lee and Chen \cite{Lee:2010}. The computational results also provided further insight into the sensitivity of the algorithm's parameters, the ability to learn to correct the choices made by the heuristic simulation policy and the added value of the proposed modifications to Monte Carlo tree search.

An interesting avenue for further work is to research the possibility of learning across different problem instances. In the proposed algorithm, the selection policy is learned in an online fashion, making it unable to benefit from information that was learned from previous problem instances. However, by using information that was learned offline (e.g. features that capture all relevant information of a node in the search space tree), the time necessary to learn a well-performing selection policy for a given problem instance could potentially be significantly reduced. It is very likely that integrating such information into the proposed algorithm would require significant changes and this problem deserves further attention.

Another interesting idea that was not further explored in this article is the ability to integrate exact solvers into the algorithm and, related to this, compare \blue{it} with other search paradigms (e.g. heuristic tree search algorithms or matheuristics). In the deeper levels of the search space tree, the number of decision variables whose values are not yet known is low, making exact approaches feasible. However, the time for a single iteration of the algorithm could potentially be significantly increased by using exact solvers. In this case, a good balance between the accuracy and time increase must be found. The question of how to do this in an optimal way is a challenging one and requires further attention. It is likely that this will also depend on the problem that one is solving and another interesting opportunity for further work is to implement the proposed algorithm for different problems than the ones presented in the current article.

Finally, another topic that was not further explored in this article is the suitability of the proposed algorithm to be parallelized. There are several possibilities to do this. The most simple option is to let the algorithm have multiple independent runs in parallel. Another option would be to run multiple simulations in parallel in the simulation phase of the algorithm. The most sophisticated option would be to combine the learned selection policies of multiple runs into a single selection policy. It is not immediately clear how to combine these policies in an optimal way and this question requires further research.

\section*{Acknowledgments}
The computational resources and services used in this work were provided by the VSC (Flemish Supercomputer Center), funded by the Research Foundation - Flanders (FWO) and the Flemish Government - department EWI. We gratefully acknowledge the support provided by the ORDinL project (FWO-SBO S007318N, Data Driven Logistics, 1/1/2018 - 31/12/2021). This research received funding from the Flemish Government under the ``Onderzoeksprogramma Artifici\"{e}le Intelligentie (AI) Vlaanderen'' programme. Pieter Leyman is a Postdoctoral Fellow of the Research Foundation - Flanders (FWO) with contract number 12P9419N. We are also grateful to Stefan R{\o}pke (Technical University of Denmark) for providing dataset A for the quay crane scheduling problem with non-crossing constraints. \blue{Editorial consultation has been provided by Luke Connolly (KU Leuven).}
	\small
	\bibliography{references}

\newpage
\section*{Appendix A: Detailed results for the quay crane scheduling problem with non-crossing constraints}

\renewcommand{\thetable}{A\arabic{table}}
\begin{table}[H] \scriptsize\centering 
	\begin{threeparttable}
		\caption{Comparison between performance of the different algorithms for dataset A. Smaller objective function values are better.}
		\label{computationalResultsComparison} 
\begin{tabular}{cccccccccccccccccccccc} \\
			\hline\noalign{\smallskip} 
			\multicolumn{1}{c}{} & \multicolumn{1}{c}{} & \multicolumn{3}{c}{\thead{CPLEX \\(Santini et al. \cite{Santini:2014})}} & \multicolumn{2}{c}{\thead{MCTS \\(this paper)}} & \multicolumn{2}{c}{\thead{EBP \\(Lee and Chen \cite{Lee:2010})}} & \multicolumn{2}{c}{\thead{SPA \\(Zhang et al. \cite{Zhang:2017})}}\\
			\cmidrule(lr){3-5}\cmidrule(lr){6-7}\cmidrule(lr){8-9}\cmidrule(lr){10-11}
			\rot{Problem instance} & \rot{Lower bound (\ref{lowerBound1})+(\ref{lowerBound2})} & \rot{Lower bound} & \rot{Value} & \rot{Time} &  \rot{Value} & \rot{Time} & \rot{Value} & \rot{Time} & \rot{Value} & \rot{Time} \\ 
			\noalign{\smallskip}
			\hline
			\noalign{\smallskip} 
			\multicolumn{1}{c}{16-4-2,893} & 724 & 726 & \textbf{726 (0.0\%)} & 2.1 s & \textbf{726 (0.0\%)} & 10.0 s & 781 (7.6\%) & $<$0.1 s & 1,073 (47.8\%) & $<$0.1 s\\
			\multicolumn{1}{c}{16-5-2,893} & 579 & 586 & \textbf{586 (0.0\%)} & 0.7 s & \textbf{586 (0.0\%)} & 10.0 s & 665 (13.5\%) & $<$0.1 s & 928 (58.4\%) & $<$0.1 s\\
			\multicolumn{1}{c}{17-4-2,941} & 736 & 741 & \textbf{741 (0.0\%)} & 19.0 s & \textbf{741 (0.0\%)} & 10.0 s & 761 (2.7\%) & $<$0.1 s & 1,135 (53.2\%) & $<$0.1 s\\
			\multicolumn{1}{c}{17-5-2,941} & 589 & 600 & \textbf{600 (0.0\%)} & 1.8 s & \textbf{600 (0.0\%)} & 10.0 s & 636 (6.0\%) & $<$0.1 s & 902 (50.3\%) & $<$0.1 s\\
			\multicolumn{1}{c}{18-4-2,856} & 714 & 720 & \textbf{720 (0.0\%)} & 17.9 s & \textbf{720 (0.0\%)} & 10.0 s & 745 (3.5\%) & $<$0.1 s & 882 (22.5\%) & $<$0.1 s\\
\end{tabular}
	\end{threeparttable}
\end{table}

\begin{table}[H] \scriptsize\centering 
\setlength{\tabcolsep}{2.5pt}
\begin{threeparttable}
		\caption{Comparison between performance of the different algorithms for dataset B. Smaller objective function values are better.}
		\label{computationalResultsComparison2} 
		\begin{tabular}{cccccccccccccccccccccc} \\
			\hline\noalign{\smallskip} 
			\multicolumn{1}{c}{} & \multicolumn{1}{c}{} & \multicolumn{3}{c}{\thead{CPLEX \\(Santini et al. \cite{Santini:2014})}} & \multicolumn{2}{c}{\thead{MCTS \\(this paper)}} & \multicolumn{2}{c}{\thead{EBP \\(Lee and Chen \cite{Lee:2010})}} & \multicolumn{2}{c}{\thead{SPA \\(Zhang et al. \cite{Zhang:2017})}}\\
			\cmidrule(lr){3-5}\cmidrule(lr){6-7}\cmidrule(lr){8-9}\cmidrule(lr){10-11}
			\rot{Problem instance} & \rot{Lower bound (\ref{lowerBound1})+(\ref{lowerBound2})} & \rot{Lower bound} & \rot{Value} & \rot{Time} &  \rot{Value} & \rot{Time} & \rot{Value} & \rot{Time} & \rot{Value} & \rot{Time} \\ 
			\noalign{\smallskip}
			\hline
			\noalign{\smallskip} 
			\multicolumn{1}{c}{200-4-33,532} & 8,383 & 8,383 & 23,852 (184.5\%) & 3,600.0 s & \textbf{8,383 (0.0\%)} & 10.0 s & 8,471 (1.0\%) & $<$0.1 s & 13,270 (58.3\%) & $<$0.1 s\\
			\multicolumn{1}{c}{200-5-33,532} & 6,707 & 6,707 & 21,433 (219.6\%) & 3,600.0 s & \textbf{6,707 (0.0\%)} & 10.0 s & 6,744 (0.6\%) & $<$0.1 s & 11,107 (65.6\%) & $<$0.1 s\\
			\multicolumn{1}{c}{200-8-33,532} & 4,192 & 4,192 & 10,294 (145.6\%) & 3,600.0 s & \textbf{4,193 (0.0\%)} & 100.0 s & 4,237 (1.1\%) & $<$0.1 s & 7,259 (73.2\%) & $<$0.1 s\\
			\multicolumn{1}{c}{200-10-33,532} & 3,354 & 3,354 & 9,884 (194.7\%) & 3,600.0 s & \textbf{3,356 (0.1\%)} & 100.0 s & 3,440 (2.6\%) & $<$0.1 s & 6,019 (79.5\%) & $<$0.1 s\\
			\multicolumn{1}{c}{250-4-42,573} & 10,644 & 10,644 & 35,550 (234.0\%) & 3,600.0 s & \textbf{10,644 (0.0\%)} & 10.0 s & 10,655 (0.1\%) & 0.1 s & 16,925 (59.0\%) & $<$0.1 s\\
\end{tabular}
	\end{threeparttable}
\end{table}

\begin{table}[H] \scriptsize\centering
\setlength{\tabcolsep}{2.5pt} 
	\begin{threeparttable}
		\caption{Dataset A: comparison between different versions of MCTS where a subset of enhancements \blue{are} omitted (versions 1-6). Smaller objective function values are better.}
		\label{detailedVersionComparisonQCSP1} 
		\begin{tabular}{cccccccccccccccccccccc} \\
			\hline\noalign{\smallskip} 
			Problem instance & MCTS Version 1 & MCTS Version 2 & MCTS Version 3 & MCTS Version 4 & MCTS Version 5 & MCTS Version 6\\ 
			\noalign{\smallskip}
			\hline
			\noalign{\smallskip}
			\multicolumn{1}{c}{Heuristic simulation policy} & \cmark & \cmark & \cmark & \cmark & \cmark & \cmark\\ 
			\multicolumn{1}{c}{Beam} & \cmark & \cmark & \cmark & \cmark & \xmark & \xmark\\  
			\multicolumn{1}{c}{Domain reduction} & \cmark & \cmark & \xmark & \xmark & \cmark & \cmark\\ 
			\multicolumn{1}{c}{Pruning subtrees by calculating bounds} & \cmark & \xmark & \cmark & \xmark & \cmark & \xmark\\ 	
			\noalign{\smallskip}
			\hline
			\noalign{\smallskip} 
			\multicolumn{1}{c}{16-4-2,893} &\textbf{726}&739&\textbf{726}&739&\textbf{726}&739\\
			\multicolumn{1}{c}{16-5-2,893} &\textbf{586}&611&\textbf{586}&611&\textbf{586}&611\\
\end{tabular}
	\end{threeparttable}
\end{table}

\begin{table}[H] \scriptsize\centering
\setlength{\tabcolsep}{2.5pt} 
	\begin{threeparttable}
		\caption{Dataset A: comparison between different versions of MCTS where a subset of enhancements \blue{are} omitted (versions 7-12). Smaller objective function values are better.}
		\label{detailedVersionComparisonQCSP2} 
		\begin{tabular}{cccccccccccccccccccccc} \\
			\hline\noalign{\smallskip} 
			Problem instance & MCTS Version 7 & MCTS Version 8 & MCTS Version 9 & MCTS Version 10 & MCTS Version 11 & MCTS Version 12\\ 
			\noalign{\smallskip}
			\hline
			\noalign{\smallskip}
			\multicolumn{1}{c}{Heuristic simulation policy} & \cmark & \cmark & \xmark & \xmark & \xmark & \xmark\\ 
			\multicolumn{1}{c}{Beam} & \xmark & \xmark & \cmark & \cmark & \cmark & \cmark\\  
			\multicolumn{1}{c}{Domain reduction} & \xmark & \xmark & \cmark & \cmark & \xmark & \xmark\\ 
			\multicolumn{1}{c}{Pruning subtrees by calculating bounds} & \cmark & \xmark & \cmark & \xmark & \cmark & \xmark\\ 	
			\noalign{\smallskip}
			\hline
			\noalign{\smallskip} 
			\multicolumn{1}{c}{16-4-2,893} &\textbf{726}&739&\textbf{726}&739&\textbf{726}&742\\
			\multicolumn{1}{c}{16-5-2,893} &\textbf{586}&611&\textbf{586}&628&\textbf{586}&625\\
\end{tabular}
	\end{threeparttable}
\end{table}

\begin{table}[H] \scriptsize\centering
\setlength{\tabcolsep}{2.5pt} 
	\begin{threeparttable}
		\caption{Dataset A: comparison between different versions of MCTS where a subset of enhancements \blue{are} omitted (versions 13-16). Smaller objective function values are better.}
		\label{detailedVersionComparisonQCSP3} 
		\begin{tabular}{cccccccccccccccccccccc} \\
			\hline\noalign{\smallskip} 
			Problem instance & MCTS Version 13 & MCTS Version 14 & MCTS Version 15 & MCTS Version 16\\ 
			\noalign{\smallskip}
			\hline
			\noalign{\smallskip}
			\multicolumn{1}{c}{Heuristic simulation policy} & \xmark & \xmark & \xmark & \xmark\\ 
			\multicolumn{1}{c}{Beam} & \xmark & \xmark & \xmark & \xmark\\  
			\multicolumn{1}{c}{Domain reduction} & \cmark & \cmark & \xmark & \xmark\\ 
			\multicolumn{1}{c}{Pruning subtrees by calculating bounds} & \cmark & \xmark & \cmark & \xmark\\ 	
			\noalign{\smallskip}
			\hline
			\noalign{\smallskip} 
			\multicolumn{1}{c}{16-4-2,893} &\textbf{726}&740&\textbf{726}&741\\
			\multicolumn{1}{c}{16-5-2,893} &\textbf{586}&616&\textbf{586}&625\\
\end{tabular}
	\end{threeparttable}
\end{table}

\begin{table}[H] \scriptsize\centering
\setlength{\tabcolsep}{2.5pt} 
	\begin{threeparttable}
		\caption{Dataset B: comparison between different versions of MCTS where a subset of enhancements \blue{are} omitted (versions 1-6). Smaller objective function values are better.}
		\label{detailedVersionComparisonQCSP4} 
		\begin{tabular}{cccccccccccccccccccccc} \\
			\hline\noalign{\smallskip} 
			Problem instance & MCTS Version 1 & MCTS Version 2 & MCTS Version 3 & MCTS Version 4 & MCTS Version 5 & MCTS Version 6\\ 
			\noalign{\smallskip}
			\hline
			\noalign{\smallskip}
			\multicolumn{1}{c}{Heuristic simulation policy} & \cmark & \cmark & \cmark & \cmark & \cmark & \cmark\\ 
			\multicolumn{1}{c}{Beam} & \cmark & \cmark & \cmark & \cmark & \xmark & \xmark\\  
			\multicolumn{1}{c}{Domain reduction} & \cmark & \cmark & \xmark & \xmark & \cmark & \cmark\\ 
			\multicolumn{1}{c}{Pruning subtrees by calculating bounds} & \cmark & \xmark & \cmark & \xmark & \cmark & \xmark\\ 	
			\noalign{\smallskip}
			\hline
			\noalign{\smallskip} 
                   \multicolumn{1}{c}{200-4-33,532} &\textbf{8,383}&\textbf{8,383}&\textbf{8,383}&\textbf{8,383}&\textbf{8,383}&\textbf{8,383}\\
			\multicolumn{1}{c}{200-5-33,532} &\textbf{6,707}&\textbf{6,707}&\textbf{6,707}&\textbf{6,707}&\textbf{6,707}&\textbf{6,707}\\
\end{tabular}
	\end{threeparttable}
\end{table}

\begin{table}[H] \scriptsize\centering
\setlength{\tabcolsep}{2.5pt} 
	\begin{threeparttable}
		\caption{Dataset B: comparison between different versions of MCTS where a subset of enhancements \blue{are} omitted (versions 7-12). Smaller objective function values are better.}
		\label{detailedVersionComparisonQCSP5} 
		\begin{tabular}{cccccccccccccccccccccc} \\
			\hline\noalign{\smallskip} 
			Problem instance & MCTS Version 7 & MCTS Version 8 & MCTS Version 9 & MCTS Version 10 & MCTS Version 11 & MCTS Version 12\\ 
			\noalign{\smallskip}
			\hline
			\noalign{\smallskip}
			\multicolumn{1}{c}{Heuristic simulation policy} & \cmark & \cmark & \xmark & \xmark & \xmark & \xmark\\ 
			\multicolumn{1}{c}{Beam} & \xmark & \xmark & \cmark & \cmark & \cmark & \cmark\\  
			\multicolumn{1}{c}{Domain reduction} & \xmark & \xmark & \cmark & \cmark & \xmark & \xmark\\ 
			\multicolumn{1}{c}{Pruning subtrees by calculating bounds} & \cmark & \xmark & \cmark & \xmark & \cmark & \xmark\\ 	
			\noalign{\smallskip}
			\hline
			\noalign{\smallskip} 
                   \multicolumn{1}{c}{200-4-33,532} &\textbf{8,383}&\textbf{8,383}&8,435&8,452&8,459&8,398\\
			\multicolumn{1}{c}{200-5-33,532} &\textbf{6,707}&\textbf{6,707}&6,771&6,826&6,779&6,797\\
\end{tabular}
	\end{threeparttable}
\end{table}

\begin{table}[H] \scriptsize\centering
\setlength{\tabcolsep}{2.5pt} 
	\begin{threeparttable}
		\caption{Dataset B: comparison between different versions of MCTS where a subset of enhancements \blue{are} omitted (versions 13-16). Smaller objective function values are better.}
		\label{detailedVersionComparisonQCSP6} 
		\begin{tabular}{cccccccccccccccccccccc} \\
			\hline\noalign{\smallskip} 
			Problem instance & MCTS Version 13 & MCTS Version 14 & MCTS Version 15 & MCTS Version 16\\ 
			\noalign{\smallskip}
			\hline
			\noalign{\smallskip}
			\multicolumn{1}{c}{Heuristic simulation policy} & \xmark & \xmark & \xmark & \xmark\\ 
			\multicolumn{1}{c}{Beam} & \xmark & \xmark & \xmark & \xmark\\  
			\multicolumn{1}{c}{Domain reduction} & \cmark & \cmark & \xmark & \xmark\\ 
			\multicolumn{1}{c}{Pruning subtrees by calculating bounds} & \cmark & \xmark & \cmark & \xmark\\ 	
			\noalign{\smallskip}
			\hline
			\noalign{\smallskip} 
                   \multicolumn{1}{c}{200-4-33,532} &8,478&8,477&8,447&8,476\\
			\multicolumn{1}{c}{200-5-33,532} &6,843&6,826&6,744&6,845\\
\end{tabular}
	\end{threeparttable}
\end{table}

\begin{table}[H] \scriptsize\centering
\setlength{\tabcolsep}{2.5pt} 
	\begin{threeparttable}
		\caption{Dataset A: comparsion between different selection policies of MCTS.}
		\label{differentSelectionPolicies2} 
		\begin{tabular}{cccccccccccccccccccccc} \\
			\hline\noalign{\smallskip} 
			Problem instance & Lower bound & \thead{MCTS\\(selection policy uses (\ref{argmaxExpression2}))} & \thead{MCTS\\(selection policy uses (\ref{argmaxExpression3}))} & \thead{OCBA-MCTS\\(Li et al. \cite{Li:2021})}\\ 
			\noalign{\smallskip}
			\hline
			\noalign{\smallskip}
			\multicolumn{1}{c}{16-4-2,893} &726&\textbf{726 (0.0\%)}&\textbf{726 (0.0\%)}&\textbf{726 (0.0\%)}\\
			\multicolumn{1}{c}{16-5-2,893} &586&\textbf{586 (0.0\%)}&\textbf{586 (0.0\%)}&\textbf{586 (0.0\%)}\\
			\multicolumn{1}{c}{17-4-2,941} &741&\textbf{741 (0.0\%)}&\textbf{741 (0.0\%)}&\textbf{741 (0.0\%)}\\
			\multicolumn{1}{c}{17-5-2,941} &600&\textbf{600 (0.0\%)}&\textbf{600 (0.0\%)}&\textbf{600 (0.0\%)}\\
			\multicolumn{1}{c}{18-4-2,856} &720&\textbf{720 (0.0\%)}&\textbf{720 (0.0\%)}&\textbf{720 (0.0\%)}\\
			\multicolumn{1}{c}{18-5-2,856} &579&\textbf{579 (0.0\%)}&\textbf{579 (0.0\%)}&\textbf{579 (0.0\%)}\\
\end{tabular}
	\end{threeparttable}
\end{table}

\begin{table}[H] \scriptsize\centering
\setlength{\tabcolsep}{2.5pt} 
	\begin{threeparttable}
		\caption{Dataset B: comparsion between different selection policies of MCTS.}
		\label{differentSelectionPolicies3} 
		\begin{tabular}{cccccccccccccccccccccc} \\
			\hline\noalign{\smallskip} 
			Problem instance & Lower bound & \thead{MCTS\\(selection policy uses (\ref{argmaxExpression2}))} & \thead{MCTS\\(selection policy uses (\ref{argmaxExpression3}))} & \thead{OCBA-MCTS\\(Li et al. \cite{Li:2021})}\\ 
			\noalign{\smallskip}
			\hline
			\noalign{\smallskip}
                   \multicolumn{1}{c}{200-4-33,532} &8,383 &\textbf{8,383 (0.0\%)}&\textbf{8,383 (0.0\%)}&\textbf{8,383 (0.0\%)}\\
			\multicolumn{1}{c}{200-5-33,532} &6,707 &\textbf{6,707 (0.0\%)}&\textbf{6,707 (0.0\%)}&\textbf{6,707 (0.0\%)}\\
			\multicolumn{1}{c}{200-8-33,532} &4,192&\textbf{4,193 (0.0\%)}&4,194 (0.0\%)&\textbf{4,193 (0.0\%)}\\
			\multicolumn{1}{c}{200-10-33,532} &3,354&\textbf{3,356 (0.1\%)}&3,361 (0.2\%)&3,360 (0.2\%)\\
			\multicolumn{1}{c}{250-4-42,573} &10,644&\textbf{10,644 (0.0\%)}&\textbf{10,644 (0.0\%)}&\textbf{10,644 (0.0\%)}\\
			\multicolumn{1}{c}{250-5-42,573} &8,515&\textbf{8,515 (0.0\%)}&\textbf{8,515 (0.0\%)}&\textbf{8,515 (0.0\%)}\\
\end{tabular}
	\end{threeparttable}
\end{table}

\newpage
\newpage
\pagebreak
\section*{Appendix B: Detailed results for the 0-1 knapsack problem}

\renewcommand{\thetable}{B\arabic{table}}

\begin{table}[h!] \scriptsize\centering 
	\begin{threeparttable}
		\caption{Comparison between Combo and MCTS for dataset A}
		\label{ComboMCTSComparisonA} 
		\begin{tabular}{cccccccccccccccccccccc} \\
			\hline\noalign{\smallskip} 
			\multicolumn{1}{c}{} & \multicolumn{2}{c}{Combo} & \multicolumn{2}{c}{MCTS}\\
			\cmidrule(lr){2-3}\cmidrule(lr){4-5}
			\noalign{\smallskip} 
			Problem instance & Value & Time & Value & Time \\ 
			\noalign{\smallskip}
			\hline
			\noalign{\smallskip} 
\multicolumn{1}{c}{strCorrSpan-50-371,096,080} & \textbf{743,839,440} & $<$0.1 s & \textbf{743,839,440 ($\mathbf{0.0}$ \%)} & 10.0 s\\
\multicolumn{1}{c}{strCorrSpan-50-646,374,982} & \textbf{953,046,980} &  $<$0.1 s & \textbf{953,046,980 ($\mathbf{0.0}$ \%)} & 10.0 s\\
\multicolumn{1}{c}{strCorrSpan-50-2,134,353,461} & \textbf{2,527,368,539} &  $<$0.1 s & \textbf{2,527,368,539 ($\mathbf{0.0}$ \%)} & 10.0 s\\
\multicolumn{1}{c}{strCorrSpan-200-1,707,778,944} & \textbf{3,246,231,168} &  $<$0.1 s & \textbf{3,246,231,168 ($\mathbf{0.0}$ \%)} & 10.0 s\\
\multicolumn{1}{c}{strCorrSpan-200-2,817,370,945} & \textbf{4,061,278,176} &  $<$0.1 s & \textbf{4,061,278,176 ($\mathbf{0.0}$ \%)} & 10.0 s\\
\end{tabular}
	\end{threeparttable}
\end{table}

\begin{table}[h!] \scriptsize\centering 
	\begin{threeparttable}
		\caption{Comparison between Combo and MCTS for dataset B}
		\label{ComboMCTSComparisonB} 
		\begin{tabular}{cccccccccccccccccccccc} \\
			\hline\noalign{\smallskip} 
			\multicolumn{1}{c}{} & \multicolumn{2}{c}{Combo} & \multicolumn{2}{c}{MCTS}\\
			\cmidrule(lr){2-3}\cmidrule(lr){4-5}
			\noalign{\smallskip} 
			Problem instance & Value & Time & Value & Time \\ 
			\noalign{\smallskip}
			\hline
			\noalign{\smallskip} 
\multicolumn{1}{c}{exp-100-10,000,000,000} & \textbf{9,999,949,957} & 2.7 s & 9,999,478,296 ($4.7 \times  10^{-3}$ \%) & 10.0 s\\
\multicolumn{1}{c}{exp-125-10,000,000,000} & \textbf{9,999,952,568} & 5.8 s & 9,999,479,807 ($4.7 \times  10^{-3}$ \%) & 10.0 s\\
\multicolumn{1}{c}{exp-150-10,000,000,000} & \textbf{9,999,954,224} & 10.5 s & 9,999,951,727 ($2.5 \times  10^{-5}$ \%) & 100.0 s\\
\multicolumn{1}{c}{exp-175-10,000,000,000} & \textbf{9,999,955,429} & 12.1 s & 9,999,953,940 ($1.5 \times  10^{-5}$ \%) & 10.0 s\\
\multicolumn{1}{c}{exp-200-10,000,000,000} & \textbf{9,999,958,144} & 18.4 s & 9,999,956,638 ($1.5 \times  10^{-5}$ \%) & 10.0 s\\
\end{tabular}
	\end{threeparttable}
\end{table}

\begin{table}[h!] \scriptsize\centering 
	\begin{threeparttable}
		\caption{Dataset A: comparison between different versions of MCTS where the modifications that we propose are introduced one by one}
		\label{detailedVersionComparisonKnapsack1} 
		\begin{tabular}{cccccccccccccccccccccc} \\
			\hline\noalign{\smallskip} 
			Problem instance & MCTS Version 1 & MCTS Version 2 & MCTS Version 3 & MCTS Version 4 & MCTS Version 5\\ 
			\noalign{\smallskip}
			\hline
			\noalign{\smallskip} 
			\multicolumn{1}{c}{Heuristic simulation policy}& \cmark& \cmark& \cmark& \cmark& \xmark\\
			\multicolumn{1}{c}{Beam}& \cmark& \cmark& \cmark& \xmark& \xmark\\
			\multicolumn{1}{c}{Domain reduction}& \cmark& \cmark& \xmark& \xmark& \xmark\\
			\multicolumn{1}{c}{Pruning subtrees by calculating bounds}& \cmark\ & \xmark& \xmark& \xmark& \xmark\\
			\noalign{\smallskip}
			\hline
			\noalign{\smallskip}
\multicolumn{1}{c}{strCorrSpan-50-371,096,080}&\textbf{743,839,440}&\textbf{743,839,440}&\textbf{743,839,440}&\textbf{743,839,440}&729,907,992\\
\multicolumn{1}{c}{strCorrSpan-50-646,374,982}&\textbf{953,046,980}&\textbf{953,046,980}&\textbf{953,046,980}&\textbf{953,046,980}&0\\
\end{tabular}
	\end{threeparttable}
\end{table}

\begin{table}[h!] \scriptsize\centering 
	\begin{threeparttable}
		\caption{Dataset B: comparison between different versions of MCTS where the modifications that we propose are introduced one by one}
		\label{detailedVersionComparisonKnapsack2} 
		\begin{tabular}{cccccccccccccccccccccc} \\
			\hline\noalign{\smallskip} 
			Problem instance & MCTS Version 1 & MCTS Version 2 & MCTS Version 3 & MCTS Version 4 & MCTS Version 5\\ 
			\noalign{\smallskip}
			\hline
			\noalign{\smallskip}
			\multicolumn{1}{c}{Heuristic simulation policy}& \cmark& \cmark& \cmark& \cmark& \xmark\\
			\multicolumn{1}{c}{Beam}& \cmark& \cmark& \cmark& \xmark& \xmark\\
			\multicolumn{1}{c}{Domain reduction}& \cmark& \cmark& \xmark& \xmark& \xmark\\
			\multicolumn{1}{c}{Pruning subtrees by calculating bounds}& \cmark\ & \xmark& \xmark& \xmark& \xmark\\
			\noalign{\smallskip}
			\hline
			\noalign{\smallskip} 
\multicolumn{1}{c}{exp-100-10,000,000,000}&\textbf{9,999,478,296}&9,999,477,943&9,999,477,490&9,995,948,086&0\\
\multicolumn{1}{c}{exp-125-10,000,000,000}&9,999,479,807&9,999,478,408&9,999,478,953&\textbf{9,999,950,674}&0\\
\end{tabular}
	\end{threeparttable}
\end{table}

\end{document}